\documentclass[letterpaper, 10 pt, conference]{ieeeconf}  
\IEEEoverridecommandlockouts                              
\usepackage{graphics} 
\usepackage{epsfig} 
\usepackage{mathptmx} 
\usepackage{times} 
\usepackage{amsmath} 
\usepackage{amssymb}  
\usepackage{caption}

\usepackage{url}            
\usepackage[table]{xcolor}
\usepackage{multirow}
\usepackage{soul}
\usepackage[flushleft,para]{threeparttable}     

\usepackage{cite}

\makeatletter
\let\NAT@parse\undefined
\makeatother

\usepackage[colorlinks=true,citecolor=green]{hyperref}

\newcommand{\se}{\mathrm{SE(3)}}

\newcommand{\variant}[1]{\texttt{#1}}

\title{\LARGE\bf Geo-Supervised Visual Depth Prediction}

\author{
Xiaohan Fei, Alex Wong, and Stefano Soatto
\thanks{This work was supported by ONR N00014-17-1-2072 and ARO W911NF-17-1-0304.}
\thanks{The authors are with the Computer Science Department, University of California, Los Angeles, USA. Email: \texttt{\small \{feixh, alexw, soatto\}@cs.ucla.edu}}
}

\begin{document}

\maketitle
\pagestyle{plain}

\begin{abstract}
We propose using global orientation from inertial measurements, and the bias it induces on the shape of objects populating the scene, to inform visual 3D reconstruction. We test the effect of using the resulting prior in depth prediction from a single image, where the normal vectors to surfaces of objects of certain classes tend to align with gravity or be orthogonal to it. Adding such a prior to baseline methods for monocular depth prediction yields improvements beyond the state-of-the-art and illustrates the power of gravity as a supervisory signal.
\end{abstract}


\section{\textsc{Introduction}}

The visual world is heavily affected by gravity, including the shape of many artifacts such as buildings and roads, and even natural objects such as trees. Gravity provides a globally consistent orientation reference that can be reliably measured with low-cost inertial sensors present in mobile devices from phones to cars. We call a machine learning system able to exploit global orientation, {\em geo-supervised}. Gravity can be easily inferred from inertial sensors without the need for dead-reckoning, and the effect of biases is negligible in the context of our application.

To measure the influence of gravity as a supervisory signal, we choose the extreme example of predicting depth from a single image. This is, literally, an impossible task in the sense that there are infinitely many three-dimensional (3D) scenes that can generate the same image. So, any process that yields a point estimate has to rely heavily on priors. We call the resulting point estimate a {\em hypothesis}, or {\em prediction}, and use public benchmark datasets to quantitatively evaluate the improvement brought about by exploiting gravity. Of course, only certain objects have a shape that is influenced by gravity.  Therefore, our prior has to be applied {\em selectively}, in a manner that is informed by the semantics of the scene. 

Our approach to geo-supervised Visual Depth Prediction is based on training a system end-to-end to produce a map from a single image and an estimate of the orientation of gravity in the (calibrated) camera frame to an inverse depth (disparity) map. In one mode of operation, the training set uses calibrated and rectified stereo pairs, together with a semantic segmentation module, to evaluate a loss function differentially on the images where geo-referenced objects are present. In a second mode, we use monocular videos instead and minimize the reprojection (prediction) error. Optionally, we can leverage modern visual-inertial odometry (VIO) and mapping systems that are becoming ubiquitous from hand-held devices to cars.

The key to our approach is a prior, or regularizer, that selectively biases certain regions of the image that correspond to geo-referenced classes such as roads, buildings, vehicles, and trees. Specifically, points in space that lie on the surface of such objects should have normals that either align with, or are orthogonal to, gravity. This is in addition to standard regularizers used for depth prediction, such as left-right consistency and piecewise smoothness.

While at training time a semantic segmentation map is needed to apply our prior selectively, it is never passed as input to the network. Therefore, at test time it is not needed, and an image is simply mapped to the disparity.

The ultimate test for a prior is whether it helps improve end-performance. 
To test our prior, we first incorporated it into two top-performing methods, one binocular (Sect.~\ref{sect-stereo}) and one monocular (Sect.~\ref{sect-mono}), in the KITTI benchmark~\cite{kitti}, and showed consistent performance improvement in all metrics. To further challenge our prior, we took two other baselines which were not the top performers. We then added our prior and tested the results against the top performers in the latest benchmark. 
We also performed generalizability tests (Sect.~\ref{sect-make3d}), ablation studies (Sect.~\ref{sect-ablation}) and demonstrated our approach with VIO on hand-held devices (Sect.~\ref{sect-visma}).

\section{\textsc{Related work}}
\label{sect-related}
Early learning-based depth prediction approaches \cite{saxena2006learning,saxena2009make3d,konrad2013learning,karsch2012depth} predict depth using local image patches and then refine  it using Markov random fields (MRFs). Recent works \cite{eigen2014depth,laina2016deeper} leverage deep networks to directly learn a representation for depth prediction where the networks are typically based on the multi-scale fully convolutional encoder-decoder structure. These methods are fully supervised and do not generalize well outside the datasets on which they are trained. Latest self-supervised methods \cite{garg2016unsupervised,godard2017unsupervised,zhou2017unsupervised} have shown better performance on benchmarks with better generalization.

There is a large body of work \cite{mahjourian2018unsupervised,yin2018geonet,wang2017learning,zhan2018unsupervised} on self-supervised monocular depth prediction following Godard \textit{et al}.~\cite{godard2017unsupervised} and Zhou \textit{et al}.~\cite{zhou2017unsupervised}, which simply use the reprojection error as a learning criterion, as has been customary in 3D reconstruction for decades. Generic priors such as piecewise smoothness and left-right consistency are also encoded into the network as additional loss terms. Our work is in-line with these self-supervised approaches, but we also exploit class-specific regularizers beyond the generic ones.

In terms of exploiting the relation of different geometric quantities in an end-to-end learning framework, closely related works include \cite{wang2016surge,qi2018geonet,liu2018planenet}, where surface normals are explicitly computed by using either a network~\cite{wang2016surge} or some heuristics~\cite{qi2018geonet}. While the former is computation intensive, the latter relies on heuristics and thus is sub-optimal. In contrast, by using losses proposed in this paper, we directly regularize depth via the depth-gravity relation without a separate surface normal predictor. Besides, both \cite{wang2016surge} and \cite{liu2018planenet} are supervised, while ours is self-supervised with the photometric loss and guided by global orientation and the semantics of the scene. 

Earlier work on semantic segmentation~\cite{shotton2008semantic} relied on local features, and have been improved by incorporating global context using various structured prediction techniques~\cite{krahenbuhl2011efficient,russell2009associative}. Starting from the work of Long \textit{et al}.~\cite{long2015fully}, fully convolutional encoder-decoder networks have been a staple in semantic segmentation. Although we do not address semantic segmentation, we leverage per-pixel semantic labeling enabled by existing systems to aid depth prediction in the form of providing class-specific priors and an attention mechanism to selectively apply such priors, which is different from joint segmentation and depth prediction approaches~\cite{jafari2017analyzing}.

The idea of using class-specific priors to facilitate reconstruction is not new \cite{hane2013joint,kundu2014joint}. In \cite{hane2013joint}, class-specific shape priors in the form of spatially varying anisotropic smoothness terms are used in an energy minimization framework to reconstruct small objects. Though promising, this system does not scale well. An efficient inference framework \cite{krahenbuhl2011efficient} has been used with a CRF model over a voxel-grid to achieve real-time performance by \cite{kundu2014joint}. While all these methods explore class-specific priors in various ways, none has used them in an end-to-end learning framework. Also, all the methods above take range images as inputs, which are then fused with semantics during optimization, while ours exploits semantics at an earlier stage --  when generating such range images which themselves can serve as priors for dense reconstruction and other inference tasks. 

\section{\textsc{Methodology}}

\begin{figure}[h]
    \centering
    \includegraphics[width=0.85\linewidth]{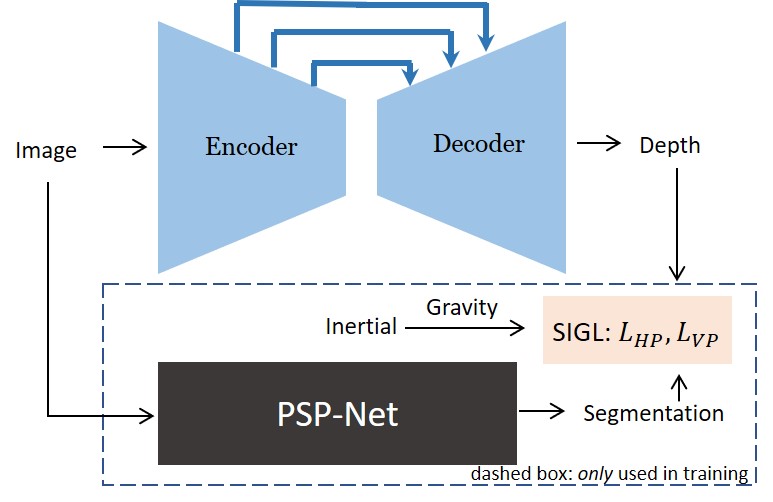}
    \caption{\small \textit{Illustration of geo-supervised visual depth prediction.} Our visual depth predictor is an encoder-decoder convolutional neural network with skip connections. At inference time, the network takes an RGB image as the only input and outputs an inverse depth map. At training time, gravity extracted from inertial measurements biases the depth prediction \textit{selectively}, which is informed by semantic segmentation produced by PSPNet. The other identical stream of the network and the photometric losses used for training are omitted in this figure.}
    \label{fig-diagram}
\end{figure}

In this section, we introduce our loss functions as regularizers added to existing models at training time, in addition to data terms (photometric loss) and generic regularizers (smoothness loss). We dub our loss semantically informed geometric loss (SIGL) because geometric constraints are selectively applied to certain image regions, where a semantic segmentation module informs the selection. Fig.~\ref{fig-diagram} illustrates part of our training diagram. In Sect.~\ref{sect-baseline}, we review baseline models used in our experiments and show that the application of our losses on top of them improves performance (Sect.~\ref{sect-exp}).

\subsection{Semantically informed geometric loss}
\label{sect-sigl}
During training, we assume to be given a partition of the image plane into semantic classes $c\in C$ that have a consistent geometric correlate. For instance, a pixel with image coordinates $(x, y) \in {\mathbb R}^2$ and class $c(x,y) =$ ``road'' is often associated to a normal plane oriented along the vertical direction (direction of gravity), whereas $c = $``building'' has a normal vector orthogonal to it. We also assume we are given the calibration matrix $K$ of the camera capturing the images, so the pixel coordinates $(x,y)$  on the image plane back-project to points in space via
\begin{equation}
{\bf X} = \left[\begin{array}{c} X \\ Y \\ Z \end{array}\right] = K^{-1} 
\begin{bmatrix}
x\\
y\\
1
\end{bmatrix}
Z(x, y)
\label{eq-back-project}
\end{equation}
where $Z(x, y)$ is the depth $Z$ of the point along the projection ray determined by $(x, y)$.

Any subset $\Omega \subset {\mathbb R}^2$ of the image plane that is the image of a spatial plane with normal vector $N \in \mathbb R^3$, at distance $\| N \|$ from the center of projection, satisfies a constraint of the form ${\bf X}^T_i N = 1$ for all $i$, assuming the plane does not go through the optical center. Stacking all the points into matrix $\bar {\bf X}\doteq [{\bf X}_1, {\bf X}_2\cdots {\bf X}_M]^\top$, we have $\bar {\bf X} N = {\bf 1}$, where ${\bf 1}$ is a vector of $M$ ones, and $M=|\Omega|$ is the cardinality of the set $\Omega$. If the direction, but not the norm, of the vector $N$ is known, a scale-invariant constraint can be easily obtained by removing the mean of the points, so that (details in Sect.~\ref{sect-derivation})
\begin{equation}
({\bf I}-\frac{1}{M}{\bf 1}{\bf 1}^\top) \bar {\bf X} N = 0.
\label{eq-constraint}
\end{equation}
The scale-invariant constraint above can be used to define a loss to penalize deviation from planarity:
\begin{equation}
 L_{HP}(\Omega_{HP}) = \frac{1}{|\Omega_{HP}|}\| ({\bf I}-\frac{1}{|\Omega_{HP}|}{\bf 1}{\bf 1}^\top)\bar {\bf X}\gamma\|_2^2
 \label{eq-lhp}
\end{equation}
where $N$ in Eq.~\eqref{eq-constraint} is replaced by normalized gravity $\gamma$ due to the homogeneity of Eq.~\eqref{eq-constraint}, and the squared norm is taken assuming the network predicts per-pixel depth $Z(x,y)$ up to additive zero-mean Gaussian noise. $\Omega_{HP} \subset \mathbb R^2$ is a subset of the image plane whose associated semantic classes have horizontal surfaces, such as ``road'', ``sidewalk'', ``parking lot'', etc. We call this loss ``horizontal plane'' loss, where the direction of gravity $\gamma$ can be reliably and globally estimated. 

Similarly, a ``vertical plane'' loss can be constructed to penalize deviation from a vertical plane whose normal $N$ has \textit{both unknown direction and norm} but lives in the null space of $\gamma$, {\em i.e.,} $N \in {\mathcal N}(\gamma)$. Thus, the vertical plane loss reads
\begin{equation}
 L_{VP}(\Omega_{VP})=\min_{\substack{N \in \mathcal N(\gamma) \\ \|N\| =1}} 
 \frac{1}{|\Omega_{VP}|}\|({\bf I}-\frac{1}{|\Omega_{VP}|}{\bf 1}{\bf 1}^\top) \bar{\bf X}N)\|_2^2
 \label{eq-lvp}
\end{equation}
where the constraint $\|N\|=1$ avoids trivial solutions $N=0$ again due to the homogeneity of the objective; $\Omega_{VP}$ is a subset of the image plane whose associated semantic classes have vertical surfaces, such as ``building'', ``fence'', ``billboard'', etc. The constrained minimization problem in the vertical plane loss $L_{VP}$ is due to the unknown direction of the surface normals and introduces some difficulties in training. We discuss approximations in Sect.~\ref{sect-derivation}.
\subsection{Explanation of the objectives}
\label{sect-derivation}
Our idea is essentially to use priors about surface normals to regularize depth prediction. An intuitive way to achieve this is to compute the surface normals from the depth values first and then impose regularity, which will eventually bias the depth predictor via backpropagation. However, such a method involves normal estimation from depth, which can be problematic, especially with a simplistic but noisy normal estimator \cite{qi2018geonet}.\footnote{For instance, one can compute the point-wise surface normal as the cross product of two vectors tangent to the surface, where the tangent vectors are approximated by connecting the underlying point to its nearest neighbors on the surface.} On the other hand, one could train a deep network to compute surface normals \cite{wang2016surge}, which is costly. Therefore, \textit{we do not compute surface normals but directly regularize the depth values} via the scale-invariant constraint Eq.~\eqref{eq-constraint} which is a function of depth and the direction of gravity. 


In what follows, we give an explanation of $L_{HP}$ Eq.~\eqref{eq-lhp} from a statistical perspective. 
Let $M=|\Omega_{HP}|$ to avoid notation clutter and expand Eq.~\eqref{eq-lhp}:
\begin{align}
& ({\bf I}-\frac{1}{M}{\bf 1}{\bf 1}^\top) \bar {\bf X} \gamma \\
= &
\begin{bmatrix}
1-\frac{1}{M} &  \cdots & -\frac{1}{M}\\
\vdots & \ddots & \vdots\\
-\frac{1}{M} &  \cdots & 1-\frac{1}{M}
\end{bmatrix}
\begin{bmatrix}
 {\bf X}_1^\top \gamma\\
 {\bf X}_2^\top \gamma\\
 \cdots\\
 {\bf X}_M^\top \gamma
\end{bmatrix}
=  
\begin{bmatrix}
\vdots\\
\big( {\bf X}_i - \frac{1}{M}\sum_{j=1}^M {\bf X}_j \big)^\top \gamma \\
\vdots
\end{bmatrix}
\end{align}
Let $\mu=\frac{1}{M}\sum_{j=1}^M \mathbf{X}_j$ be the sample mean of the 3D coordinates and the horizontal plane loss $L_{HP}$ reads
\begin{equation}
 L_{HP}(\Omega_{HP})= \frac{1}{M}\sum_{i=1}^M \big( ({\bf X}_i - \mu)^\top \gamma \big)^2
\end{equation}
which is the sample variance of the 3D coordinates projected to the direction of gravity $\gamma$ (coinciding with the surface normal for horizontal planes). To minimize $L_{HP}$ is to minimize the variance of the 3D coordinates along the surface normal.

Similarly, to minimize $L_{VP}$ Eq.~\eqref{eq-lvp} is to minimize the variance of the 3D coordinates along some direction perpendicular to gravity. However, if the direction is unknown, one needs to jointly solve the direction while minimizing $L_{VP}$, which explains the constrained quadratic problem in $L_{VP}$. Though this can be solved via eigendecomposition, the gradients of the solver -- needed in backpropagation -- are non-trivial to compute. In fact, representing an optimization procedure as a layer of a neural network is an open research problem \cite{amos2017optnet}. To alleviate both numerical and implementation difficulties, we uniformly sample unit vectors from the null space of gravity and compute the minimum of the objective over the samples as an approximation to the loss. Empirically, we found using eight directions sampled every 45 degrees from 0 to 360 generally performs well.

\subsection{View synthesis as supervision and baselines}
\label{sect-baseline}
To showcase the ability to improve upon existing self-supervised monocular depth prediction networks, we add our losses to two publicly available models -- Godard~\cite{godard2017unsupervised} (\variant{LR-Consistency}) and Yin~\cite{yin2018geonet} (\variant{GeoNet}) -- as baselines and perform both quantitative and qualitative comparisons. We additionally apply our losses to Zhan ~\cite{zhan2018unsupervised} (\variant{Stereo-Temporal}) and Wang~\cite{wang2017learning} (\variant{DDVO}), the state-of-the-art methods in their respective training setting, stereo pairs/videos, and monocular videos. \variant{LR-Consistency} is trained with rectified stereo image pairs,  \variant{GeoNet} and \variant{DDVO} use monocular videos while \variant{Stereo-Temporal} uses stereo videos. At test time, all training settings result in a system that takes a single image as input and predicts an inverse depth map as output. We show that by applying our losses to the baselines \variant{LR-Consistency} and \variant{GeoNet}, we achieve better performance than the state-of-the-art methods \variant{Stereo-Temporal} and \variant{DDVO}. Furthermore, we produce new state-of-the-art results by applying our losses to \variant{Stereo-Temporal} and \variant{DDVO}.

\subsubsection{Training with stereo pairs}
\label{sect-method-stereo}
At training time, our first baseline model (\variant{LR-Consistency}) takes a single left image as its input and predicts two disparity maps $D^L, D^R: \mathbb{R}^2 \supset \Omega \rightarrow \mathbb{R}_+$ for both left and right cameras. The network follows the fully convolutional encoder-decoder structure with skip connections. The total loss consists of three terms: Appearance loss, smoothness of disparity and left-right consistency, each of which is evaluated on both the left and the right streams across multiple scale levels. Here we address the view synthesis loss, which serves as the data term and is part of the appearance loss:
\begin{align}
L_\text{vs}^L= \frac{1}{|\Omega|} \sum_{(x,y) \in \Omega} \| I^L(x,y) - I^R(x+D^L(x, y), y) \|_1.
\label{eq-stereo}
\end{align}
The view synthesis loss is essentially the photometric difference of the left image $I^L(x, y)$ and the right image warped to the left view $I^R(x+D^L(x, y), y)$ according to the left disparity prediction $D^L(x, y)$. The right view synthesis loss is constructed in the same way. Though only one disparity map is needed at inference time, it has been shown that predicting both left and right disparity maps and including the left-right consistency loss Eq.~\eqref{eq-consistency} are in general beneficial~\cite{godard2017unsupervised}.

\begin{equation}
L_\text{lr}^L = \frac{1}{|\Omega|} \sum_{(x,y) \in \Omega} \| D^L(x,y) - D^R(x+D^L(x, y), y)\|_1
\label{eq-consistency}
\end{equation}

\subsubsection{Training with stereo videos}
\label{sect-stereo-temporal}
In our second baseline \variant{Stereo-Temporal}, stereo videos are used to train a monocular depth predictor, where two frames of a stereo pair and another frame one time step ahead are involved in constructing a stereo-temporal version of the photometric loss: For the stereo pair, Eq.~\eqref{eq-stereo} is applied while for the temporal pair, Eq.~\eqref{eq-mono} (detailed below) is applied.

\subsubsection{Training with monocular videos}
\label{sect-method-mono}
To train our third and fourth baseline models (\variant{GeoNet} and \variant{DDVO}), a single reference frame $I_t$ is fed into the depth network and frames $I_{t'}, t' \in W_t$ in a temporal window centered at $t$ are used to construct the view synthesis loss, also known as reprojection error:
\begin{equation}
L_\text{vs}=\frac{1}{|W_t||\Omega|}
\sum_{t' \in W_t} \sum_{(x,y) \in \Omega}
\| I_{t}(x, y)-I_{t'}\big(\pi(\hat g_{t't} {\bf X})\big)\|_1
\label{eq-mono}
\end{equation}
which is the difference between the reference frame $I_t$ and neighboring frames $I_{t'}$ warped to it. ${\bf X}$ is the back-projected point defined in Eq.~\eqref{eq-back-project}, $\pi$ is a central (perspective) projection, and $\hat g_{t't}$ is the relative camera pose up to an unknown scale predicted by an auxiliary pose network which takes both $I_t$ and $I_{t'}$ as its input. Note that the pose and depth networks are coupled via the view synthesis loss at training time; at test time, the depth network alone is needed to perform depth prediction with a single image as its input. Interestingly, in Sect.~\ref{sect-visma} we found that replacing the pose network with pose estimation from VIO produces better results compared to the multi-task learning diagram where pose and depth networks are trained simultaneously, which sheds light on the use of classic SLAM/Odometry systems in developing better learning algorithms.

A detailed discussion about other losses serving as regularization terms is beyond the scope of this paper and can be found in \cite{godard2017unsupervised,zhou2017unsupervised,yin2018geonet,wang2017learning}. 

\section{\textsc{Implementation Details}}
\label{sect-impl}
\subsection{Semantic segmentation}
\label{sect-pspnet}
At training time, we use PSPNet~\cite{zhao2017pyramid} pre-trained on the CityScapes dataset~\cite{cordts2016cityscapes} provided by the authors to obtain per-pixel labeling. For every pixel $(x, y)\in \mathbb R^2$, a probability distribution over $19$ classes is predicted by PSPNet, of which the most likely class $c(x,y) \in C$ determines the orientation of the surface where the back-projected point ${\bf X}$ sits. We group the 19 classes into 7 categories\footnote{``flat'': road, sidewalk; ``human'': rider, person; ``vehicle'': car, truck, bus, train, motorcycle, bicycle; ``construction'': building, wall, fences; ``object'': pole, traffic light, traffic sign; ``nature'': vegetation, terrain; ``sky'': sky.} according to the CityScapes benchmark and test our losses on all of them. Empirically, we found that it is most beneficial to apply our losses to the ``flat'', ``vehicle'' and ``construction'' categories and therefore all the comparisons on KITTI against baseline methods are made with these categories regularized. The influence of other categories is studied in Sect.~\ref{sect-ablation}.


\subsection{Gravity}
\label{sect-gravity}
For imagery captured by a static platform equipped with an inertial measurement unit (IMU), one can use the gravity $\gamma_b \in \mathbb{R}^3$ measured in the body frame (coinciding with the IMU frame) and simply apply the body-to-camera rotation $R_{cb} \in \mathrm{SO}(3)$ to obtain the gravity in the camera frame $\gamma=R_{cb}\gamma_b$ which is then used in Eq.~\eqref{eq-lhp} and \eqref{eq-lvp}. For moving platforms, one resorts to robust VIO, which is well studied \cite{mourikis2007multi,tsotsos2015robust}. In Sect.~\ref{sect-visma}, we demonstrated our approach on a visual-inertial odometry dataset, where both camera pose and gravity are estimated online by VIO.

For our experiments on the KITTI dataset, thanks to the GPS/IMU sensor package which provides linear acceleration of the sensor platform measured both in the body frame ($\alpha_b \in \mathbb{R}^3$) and the spatial frame ($\alpha_s \in \mathbb{R}^3$), we are able to compute the spatial-to-body rotation $R_{bs} \in \mathrm{SO}(3)$ and then bring the gravity $\gamma_s=[0,0,9.8]^\top$ from the spatial frame to the camera frame $\gamma=R_{cb}R_{bs}\gamma_s$. In all settings, $R_{cb}$ (rotational part of the body-to-camera transformation) is obtained via offline calibration procedures. 

\subsection{Training details}
A GTX 1080 Ti GPU and Adam~\cite{kingma2014adam} optimizer are used in our experiments. Depending on different model variants and input image sizes, training time varies from 8 hours to 16 hours. For \variant{LR-Consistency} and \variant{GeoNet} which were initially implemented in TensorFlow, we implemented our losses also in TensorFlow and applied them to the existing code bases. Code of \variant{Stereo-Temporal} is available online, but in Caffe, thus we migrated their model to TensorFlow and applied our losses. We also implemented our losses in PyTorch, which were then applied to \variant{DDVO} of which the PyTorch version was made available by the author. Our code is available at \url{https://github.com/feixh/GeoSup}.


\section{\textsc{Experiments}}
\label{sect-exp}
To enable quantitative evaluation, we exploit the KITTI benchmark, and test our approach against the state-of-the-art as described in detail below (Sect.~\ref{sect-stereo}\&\ref{sect-mono}). We also carried out ablation studies (Sect.~\ref{sect-ablation}) and tested the generalizability of our approach (Sect.~\ref{sect-make3d}).
In addition to KITTI, which features planar motion in driving scenarios, we have conducted experiments on VISMA dataset~\cite{fei2018visual} -- an indoor visual-inertial odometry dataset captured under non-trivial ego-motion (Sect.~\ref{sect-visma}).

\subsection{KITTI Eigen split}
We compare our approach with recent state-of-the-art methods on the monocular depth prediction task using the KITTI Eigen split~\cite{eigen2014depth} in two training domains: stereo pairs/videos and monocular videos (Sect.~\ref{sect-baseline}). The Eigen split test set contains 697 test images selected from 29 of 61 scenes provided by the raw KITTI dataset.  Of the remaining 32 scenes containing 23,488 stereo pairs, 22,600 pairs are used for training, and the rest is used for validation per the training split proposed by \cite{garg2016unsupervised}. To generate ground truth depth maps for validation and evaluation, we take the Velodyne data points associated with each image and project them from the Velodyne frame to the left RGB camera frame. 
Each resulting ground truth depth map covers approximately $5\%$ of the corresponding image and may be erroneous. To handle this, first, we use the cropping scheme proposed by \cite{garg2016unsupervised}, which masks out the potentially erroneous extremities from the left, right and top areas of the ground truth depth map. Then we evaluate depth prediction only at pixels where ground truth depth is available. For visualization, we linearly interpolate each sparse depth map to cover the entire image (Fig.~\ref{fig-kitti}). 

We additionally provide quantitative evaluations of variants of the models pre-trained on CityScapes and fine-tuned on KITTI. CityScapes dataset contains 22,973 training stereo pairs captured in various cities across Germany with a similar modality as KITTI. We cropped each input image to keep only the top 80\% of the image, removing the reflective hood.

The error and accuracy metrics, which are initially proposed by \cite{eigen2014depth} and adopted by others, are used (Table~\ref{tab-metric}). Also as a convention in the literature, performances evaluated with depth prediction capped at 50 and 80 meters are reported as suggested by \cite{godard2017unsupervised}. The choice of 80 meters is two-fold: 1) maximum depth present in the KITTI dataset is on the order of 80 meters and 2) non-thresholded measures can be sensitive to the significant errors in depth caused by prediction errors at small disparity values. For the same reason, depth prediction is capped at 70 meters in the Make3D experiment. Prediction capped at 50 meters is also evaluated since depth at closer range is more applicable to real-world scenarios.

\begin{table}
    \caption{Error and Accuracy Metrics}
    \label{tab-metric}
    \centering
    \begin{threeparttable}
    \begin{tabular}{l|l}
        \hline
        Metric & Definition \\
        \hline \hline
        AbsRel & $\frac{1}{|\Omega|}\sum_{(x,y) \in \Omega} \frac{|Z(x,y)-Z^\text{gt}(x,y)|}{Z^\text{gt}(x,y)}$\\
        \hline
        SqRel & $\frac{1}{|\Omega|}\sum_{(x,y)\in\Omega}\frac{|Z(x,y)- Z^\text{gt}(x,y)|^2}{Z^\text{gt}(x,y)}$\\
        \hline
        RMSE & $\sqrt{\frac{1}{|\Omega|} \sum_{(x,y) \in \Omega}|Z(x,y) - Z^\text{gt}(x,y) |^2}$\\
        \hline
        RMSE log & $\sqrt{\frac{1}{|\Omega|} \sum_{(x,y) \in \Omega}|\log Z(x,y) - \log Z^\text{gt}(x,y)|^2}$\\
        \hline
        $\log_{10}$ & $\frac{1}{|\Omega|}\sum_{(x,y)\in\Omega}|\log Z(x,y)-\log Z^\text{gt}(x,y)|$\\
        \hline
        Accuracy & \% of $Z(x,y)$ s.t. $\delta \doteq \max\big(\frac{Z(x,y)}{Z^\text{gt}(x,y)}, \frac{Z^\text{gt}(x,y)}{Z(x,y)}\big) < \text{threshold}$ \\
        \hline
    \end{tabular}
    \begin{tablenotes}
        $Z(x,y)$ is the predicted depth at $(x,y) \in \Omega$ and $Z^\text{gt}(z,y)$ is the corresponding ground truth. Three different thresholds ($1.25, 1.25^2$ and $1.25^3$) are used in the accuracy metric as a convention in the literature.
    \end{tablenotes}
\end{threeparttable}
\vspace{-10pt}
\end{table}

\subsection{Training with stereo pairs}
\label{sect-stereo}
The first baseline we adopt is Godard~\cite{godard2017unsupervised} (with \variant{VGG}~\cite{simonyan2014very} as feature extractor), to which SIGL is imposed at training time along with the view synthesis loss Eq.~\eqref{eq-stereo} and other generic regularizers used in \cite{godard2017unsupervised}. The model is trained from scratch with stereo pairs following the Eigen split and compared to both supervised \cite{eigen2014depth,liu2016learning} and self-supervised methods \cite{godard2017unsupervised,zhan2018unsupervised}. 
In addition, we apply our losses to variants of the baseline (with \variant{ResNet}~\cite{he2016deep} as feature extractor; w/ \& w/o post-processing) and evaluate different training schemes (w/ \& w/o pre-training on CityScapes). Quantitative comparisons can be found in Table~\ref{tab-stereo}, where the results with SIGL added as an additional regularizer follow the results of the baseline models and variants. In the column marked ``Data'', \texttt{K} refers to Eigen split benchmark on the KITTI dataset, and \texttt{CS} refers to the CityScapes dataset. Methods marked with \texttt{CS+K} are pre-trained on CityScapes and then fine-tuned on KITTI Eigen split. \texttt{pp} denotes post-processing. Cap $X$m means depth predictions are capped at $X$ meters. Results of Zhan~\cite{zhan2018unsupervised} \variant{Stereo-Temporal} are taken from their paper. The rest of the results are taken from \cite{godard2017unsupervised} unless otherwise stated.

We want to remind the reader that the first baseline model atop which we built ours is Godard~\cite{godard2017unsupervised} \variant{VGG} which initially performed worse than the \variant{Stereo-Temporal} model of Zhan~\cite{zhan2018unsupervised} by a large margin, but by applying our losses to the baseline at training time we managed to boost its performance and make it perform even better than the \variant{Stereo-Temporal} model at test time. Note that the \variant{Stereo-Temporal} model also exploits temporal information in addition to stereo pairs for training while our first baseline built atop Godard does not.

As a second baseline, we apply our losses additionally to the \variant{Stereo-Temporal} model of Zhan to further push the state-of-the-art. Table~\ref{tab-stereo} shows that our losses improve the \variant{Stereo-Temporal} model across all error metrics with the accuracy metrics $\delta < 1.25^2$ and $\delta < 1.25^3$ being comparable. Another variant of Zhan's model pre-trains on NYU-V2~\cite{silberman2012nyuv2} in a fully supervised fashion and is therefore not pertinent to this comparison. 
Fig.~\ref{fig-kitti} shows a head-to-head qualitative comparison of ours and the baseline models.

\setlength{\tabcolsep}{2pt}
\begin{table}[h]
\caption{Training with stereo pairs on KITTI.}
\centering
\label{tab-stereo}
\begin{scriptsize}
\begin{threeparttable}[c]
    \begin{tabular}{l|c|cccc|ccc}
    \hline 
    Method & Data & \multicolumn{4}{c|}{Error metric} & \multicolumn{3}{c}{Accuracy $(\delta < )$} \\
    
        & & {\tiny AbsRel} & {\tiny SqRel} & {\tiny RMSE} & {\tiny RMSElog}  & $1.25$ & $1.25^2$ & $1.25^3$ \\
    \hline
    \hline
    \multicolumn{9}{c}{Depth: cap 80m}\\
    \hline
    TrainSetMean\tnote{*} & \texttt{K} & 0.361 & 4.826 & 8.102 & 0.377 & 0.638 & 0.804 & 0.894 \\
    
    Eigen~\cite{eigen2014depth} {\tiny \texttt{Coarse}}\tnote{*} & \texttt{K} & 0.214 & 1.605 & 6.563 & 0.292 & 0.673 & 0.884 & 0.957 \\
    
    Eigen~\cite{eigen2014depth} {\tiny \texttt{Fine}}\tnote{*} & \texttt{K} & 0.203 & 1.548 & 6.307 & 0.282 & 0.702 & 0.890 & 0.958 \\
    
    Liu~\cite{liu2016learning}\tnote{*} & \texttt{K} & 0.201 & 1.584 & 6.471 & 0.273 & 0.680 & 0.898 & 0.967 \\
    
    \hline
    Godard~\cite{godard2017unsupervised} {\tiny \texttt{VGG}} & \texttt{K} & 0.148 & 1.344 & 5.927 & 0.247 & 0.803 & 0.922 & 0.964\\
    
    +SIGL & \texttt{K} &  {\bf 0.139} &  {\bf 1.211} &  {\bf 5.702} &   {\bf 0.239} &   {\bf 0.816} &  {\bf 0.928} & {\bf 0.966} \\
    
    \hline
    Zhan~\cite{zhan2018unsupervised} {\tiny \texttt{Stereo-Temporal}} & \texttt{K} & 0.144 & 1.391 & 5.869 & 0.241 & 0.803 &  0.928 &  0.969 \\
    
    +SIGL & \texttt{K} & {\bf 0.137} & {\bf 1.061} & {\bf 5.692} & {\bf 0.239} & {\bf 0.805} & 0.928 & 0.969 \\
    
    
    \hline
    Godard~\cite{godard2017unsupervised} {\tiny \texttt{VGG pp}} & \texttt{CS+K} & 0.124 & 1.076 & 5.311 & 0.219 & 0.847 & 0.942 & 0.973\\
    
    +SIGL & \texttt{CS+K} &  {\bf 0.114} & {\bf 0.885} & {\bf 4.877} & {\bf 0.203} & {\bf 0.858} & {\bf 0.950} & {\bf 0.978} \\
    
    \hline
    Godard~\cite{godard2017unsupervised} {\tiny \texttt{ResNet pp}} & \texttt{CS+K} & 0.114 & 0.898 & 4.935 & 0.206 & 0.861 & 0.949 & 0.976\\
    
    +SIGL & \texttt{CS+K}  &  {\bf 0.112} & {\bf 0.836} & {\bf 4.892} & {\bf 0.204} & {\bf 0.862} &  {\bf 0.950} & {\bf 0.977} \\
    
    \hline
    \hline

    \multicolumn{9}{c}{Depth: cap 50m}\\
    \hline
    Garg~\cite{garg2016unsupervised} & \texttt{K} & 0.169 & 1.080 & 5.104 & 0.273 & 0.740 & 0.904 & 0.962 \\
    
    \hline
    Godard~\cite{godard2017unsupervised} {\tiny \texttt{VGG}} & \texttt{K} & 0.140 & 0.976 & 4.471 & 0.232 & 0.818 & 0.931 & 0.969 \\
    
    +SIGL & \texttt{K} & {\bf 0.132} & {\bf 0.891} & {\bf 4.312} & {\bf 0.225} & {\bf 0.831} & {\bf 0.936} & {\bf 0.970} \\
    
    \hline
    Zhan~\cite{zhan2018unsupervised} {\tiny \texttt{Stereo-Temporal}} & \texttt{K} & 0.135 & 0.905 &  4.366 & 0.225 & 0.818 & 0.937 & 0.973 \\
    
    +SIGL & \texttt{K} & {\bf 0.131} & {\bf 0.829} & {\bf 4.217} & {\bf 0.224} & {\bf 0.824} & 0.937 & 0.973 \\
    
    
    
    \hline
    Godard~\cite{godard2017unsupervised} {\tiny \texttt{VGG pp}} & \texttt{CS+K} & 0.112 & 0.680 &  3.810 & 0.198 & 0.866 & 0.953 & 0.979 \\
    
    +SIGL & \texttt{CS+K} &  {\bf 0.108} & {\bf 0.658} & {\bf 3.728} & {\bf 0.192} & {\bf 0.870} & {\bf 0.955} & {\bf 0.981} \\
    
    \hline
    Godard~\cite{godard2017unsupervised} {\tiny \texttt{ResNet pp}} & \texttt{CS+K} & 0.108 & 0.657 & 3.729 & 0.194 & 0.873 & 0.954 & 0.979 \\
    
    +SIGL & \texttt{CS+K} &  {\bf 0.106} & {\bf 0.615} & {\bf 3.697} & {\bf 0.192} & {\bf 0.874} & {\bf 0.956} & {\bf 0.980} \\
    
    \hline
    \end{tabular}
\begin{tablenotes}
    \item [*] With ground truth depth supervision.\\
    \item +SIGL: training with SIGL enabled
\end{tablenotes}
\end{threeparttable}
\end{scriptsize}
\end{table}
\subsection{Training with monocular videos}
\label{sect-mono}
To demonstrate the effectiveness of our loss in the second training setting (monocular videos), we impose SIGL to our third (Yin~\cite{yin2018geonet}) and fourth (Wang~\cite{wang2017learning}) baseline. Using the KITTI Eigen split, we follow the training and validation 3-frame sequence selection proposed by \cite{zhou2017unsupervised} where the first and third frames are treated as the source views and the central (second) frame is treated as the reference as in Eq.~\eqref{eq-mono}. Of the 44,540 total sequences, 40,109 are used for training and 4,431 for validation. We evaluate our system on the aforementioned 697 test images \cite{eigen2014depth}. The same training and evaluation scheme are also applied to other top-performing methods \cite{zhou2017unsupervised,mahjourian2018unsupervised} in addition to the selected baselines.

\begin{figure*}
\centering
\begin{tabular}{ccccc}
    \includegraphics[width=0.23\linewidth]{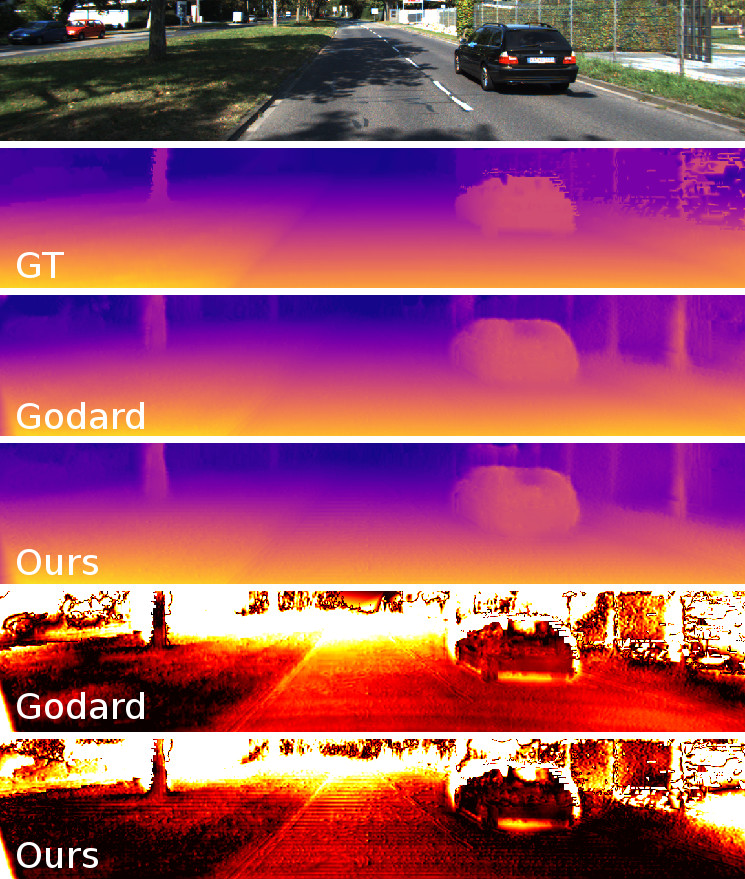} &
    \includegraphics[width=0.23\linewidth]{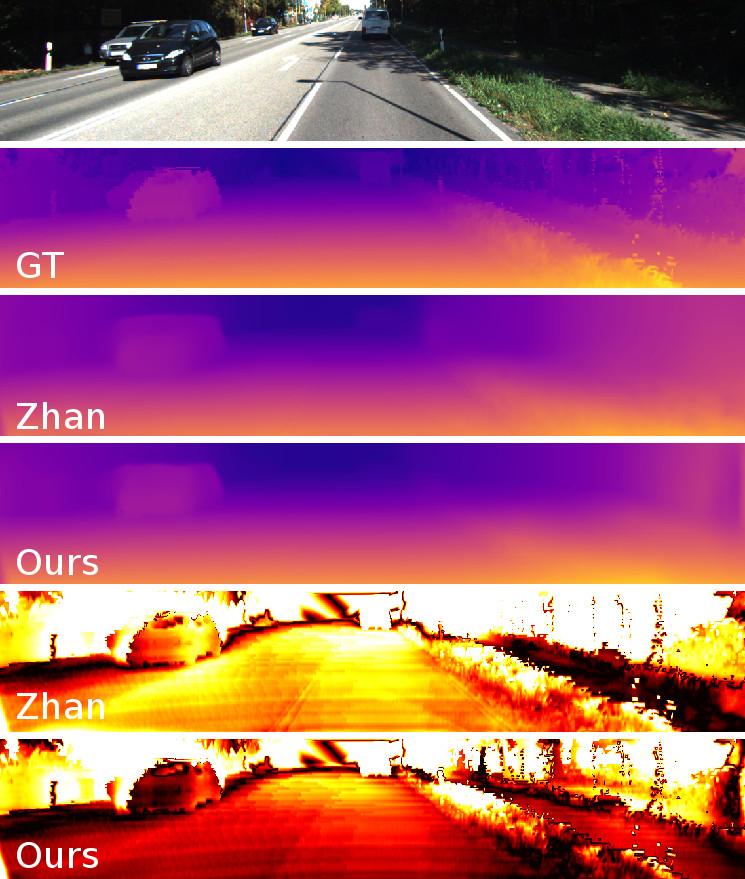} &
    \includegraphics[width=0.23\linewidth]{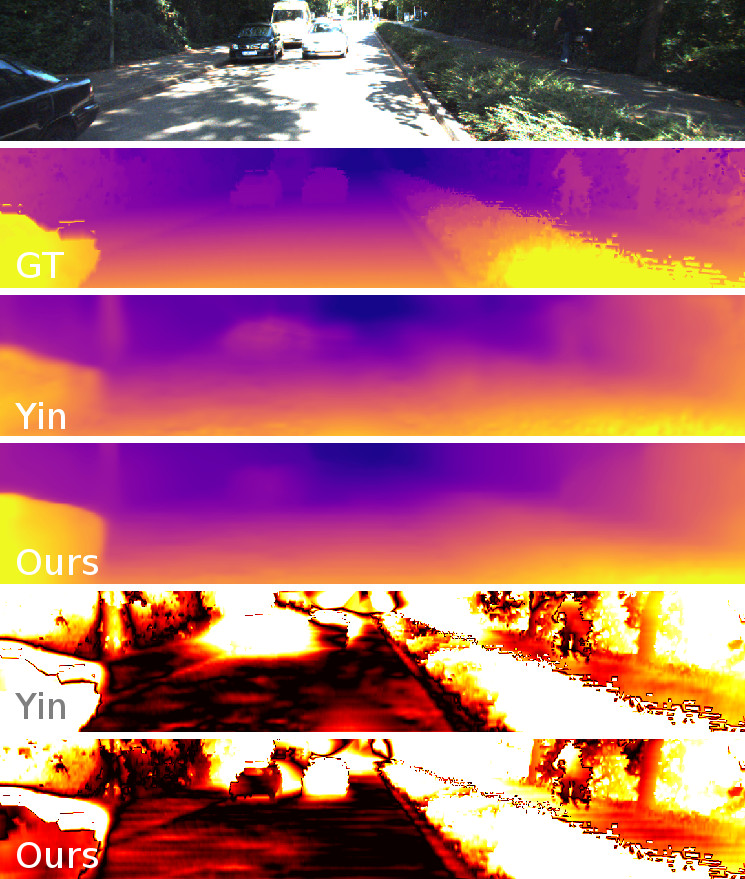} &
    \includegraphics[width=0.23\linewidth]{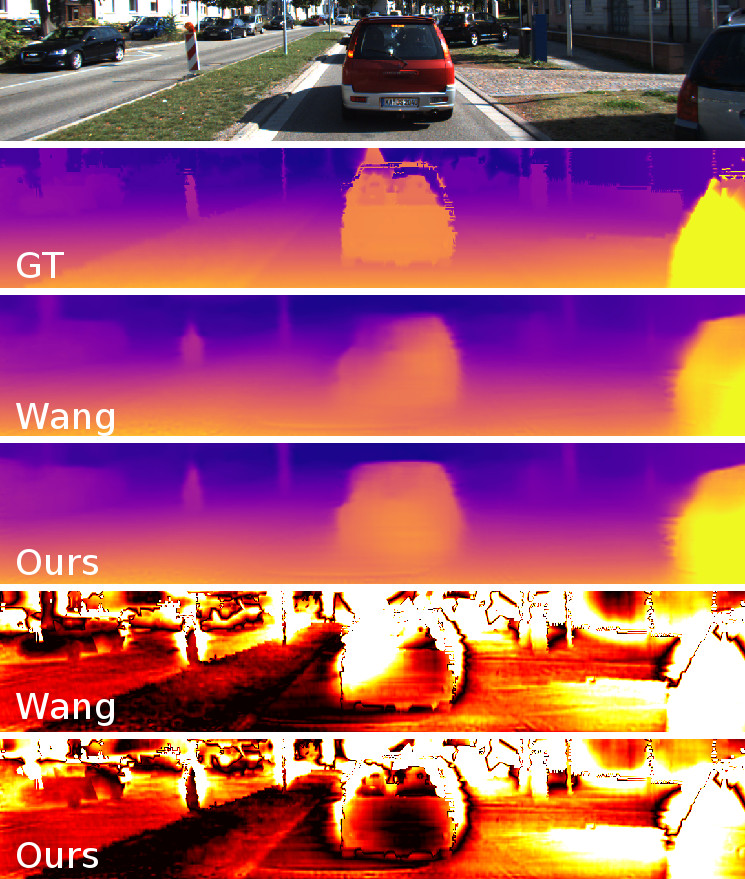} &
    \includegraphics[width=0.014\linewidth]{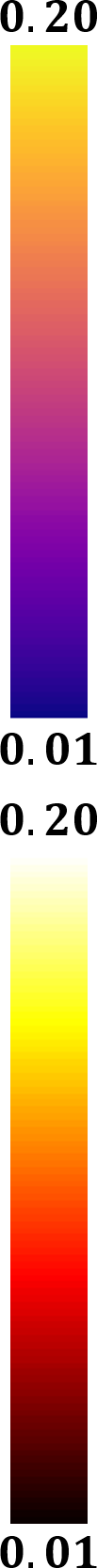}
\end{tabular}
\caption{\small \textit{Qualitative results on KITTI Eigen split.} (best viewed at $5\times$ with color) Top to bottom, each column shows an input RGB image, the corresponding ground truth inverse depth map, the predictions of baseline models trained without and with our priors, AbsRel error maps of baseline models trained without and with our priors. All the models are trained on KITTI Eigen split. For the purpose of visualization, ground truth is interpolated and all the images are cropped according to \cite{garg2016unsupervised}. For the error map, darker means smaller error. Typical image regions where we do better (darker in the error map) include cars, roads and walls.}
\label{fig-kitti}
\end{figure*}

Table~\ref{tab-mono} shows detailed comparisons against state-of-the-art self-supervised methods trained using monocular video sequences.
We compare against best-performing model variants of Wang~\cite{wang2017learning} (\variant{PoseCNN} \& \variant{PoseCNN+DDVO}) and Yin~\cite{yin2018geonet} (\variant{ResNet}) with and without pre-training on CityScapes. By adding our losses to existing models, we observe systematic performance improvement across all metrics.
Though initially performing worse than Wang~\cite{wang2017learning} \variant{PoseCNN+DDVO}, Yin~\cite{yin2018geonet} \variant{ResNet} with the proposed losses even outperforms the original \variant{PoseCNN+DDVO}. Moreover, we achieve new state-of-the-art by adding our losses to \variant{PoseCNN+DDVO} trained on both CityScapes and KITTI. Fig.~\ref{fig-kitti} illustrates representative image regions where we do better.

\setlength{\tabcolsep}{2pt}
\begin{table}[h]
\caption{Training with monocular videos on KITTI.} 
\label{tab-mono}
\centering
\begin{scriptsize}
\begin{threeparttable}[c]
    \begin{tabular}{l|c|cccc|ccc}
    \hline 
    Method & Data & \multicolumn{4}{c|}{Error metric} & \multicolumn{3}{c}{Accuracy $(\delta < )$ } \\ 
        & & {\tiny AbsRel} & {\tiny SqRel} & {\tiny RMSE} & {\tiny RMSElog} & $1.25$ & $1.25^2$ & $1.25^3$ \\
    \hline
    \hline
    \multicolumn{9}{c}{Depth: cap 80m}\\
    \hline
    
    Zhou~\cite{zhou2017unsupervised} & \texttt{K} & 0.208 & 1.768 & 6.856 & 0.283 & 0.678 & 0.885 & 0.957 \\
    
    
    Mahjourian~\cite{mahjourian2018unsupervised} & \texttt{K} & 0.163 & 1.240 & 6.220 & 0.250 & 0.762 & 0.916 & 0.968 \\
    
    Yin~\cite{yin2018geonet} {\tiny \texttt{ResNet}} & \texttt{K} & 0.155 & 1.296 & 5.857 & 0.233 & 0.793 & 0.931 & 0.973 \\
    
    +SIGL  & \texttt{K} & {\bf 0.142} & {\bf 1.124} & {\bf 5.611} & {\bf 0.223} & {\bf 0.813} & {\bf 0.938} & {\bf 0.975} \\
    
    \hline
    Wang~\cite{wang2017learning} {\tiny \texttt{PoseCNN}} & \texttt{K} & 0.155 & 1.193 & {\bf 5.613} & 0.229 & 0.797 & 0.935 & 0.975 \\
    
    +SIGL & \texttt{K} & {\bf 0.147} & {\bf 1.076} & 5.640 & {\bf 0.227} & {\bf 0.801} & 0.935 & 0.975 \\
    
    \hline
    Wang~\cite{wang2017learning} {\tiny \texttt{PoseCNN+DDVO}} & \texttt{K} & 0.151 & 1.257 & 5.583 & 0.228 & {\bf 0.810} & 0.936 & 0.974 \\
    
    +SIGL & \texttt{K} & {\bf 0.146} & {\bf 1.068} & {\bf 5.538} & {\bf 0.224} & 0.809 & {\bf 0.938} & {\bf 0.975} \\
    
    \hline
    Zhou~\cite{zhou2017unsupervised} & \texttt{CS+K} & 0.198 & 1.836 & 6.565 & 0.275 & 0.718 & 0.901 & 0.960 \\
    
    Mahjourian~\cite{mahjourian2018unsupervised} & \texttt{CS+K} & 0.159 & 1.231 & 5.912 & 0.243 & 0.784 & 0.923 & 0.970 \\
    
    Yin~\cite{yin2018geonet} {\tiny \texttt{ResNet}} & \texttt{CS+K} & 0.153 & 1.328 & 5.737 & 0.232 & 0.802 & 0.934 & 0.972 \\
    
    +SIGL  & \texttt{CS+K} & {\bf 0.147} & {\bf 1.076} & {\bf 5.468} & {\bf 0.222} & {\bf 0.806} & {\bf 0.938} & {\bf 0.976} \\
    
    \hline
    Wang~\cite{wang2017learning} {\tiny \texttt{PoseCNN+DDVO}} & \texttt{CS+K} & 0.148 & 1.187 & 5.496 & 0.226 & 0.812 & 0.938 & 0.975 \\
    +SIGL & \texttt{CS+K} & {\bf 0.142} & {\bf 1.094} & {\bf 5.409} & {\bf 0.219} & {\bf 0.821} & {\bf 0.941} & {\bf 0.976} \\
    
    \hline
    \hline
    \multicolumn{9}{c}{Depth: cap 50m}\\
    \hline
    Zhou~\cite{zhou2017unsupervised} & \texttt{K} & 0.201 & 1.391 & 5.181 & 0.264 & 0.696 & 0.900 & 0.966 \\
    
    Mahjourian~\cite{mahjourian2018unsupervised} & \texttt{K} & 0.155 & 0.927 & 4.549 & 0.231 & 0.781 & 0.931 & 0.975 \\
    
    Yin~\cite{yin2018geonet} {\tiny \texttt{ResNet}} & \texttt{K} & 0.147 & 0.936 & 4.348 & 0.218 & 0.810 & 0.941 & 0.977 \\
    
    +SIGL  & \texttt{K} & {\bf 0.135} & {\bf 0.834} & {\bf 4.193} & {\bf 0.208} & {\bf 0.831} & {\bf 0.948} & {\bf 0.979} \\
    
    \hline
    Wang~\cite{wang2017learning} {\tiny \texttt{PoseCNN}}\tnote{$\dagger$} & \texttt{K} &  0.149 & 0.920 & 4.303 & 0.216 & 0.813 & 0.943 & 0.979 \\
    
    +SIGL & \texttt{K} & {\bf 0.140} & {\bf 0.816} & {\bf 4.234} & {\bf 0.212} & {\bf 0.818} & {\bf 0.945} & {\bf 0.980} \\
    
    \hline
    Wang~\cite{wang2017learning} {\tiny \texttt{PoseCNN+DDVO}}\tnote{$\dagger$} & \texttt{K} & 0.144 & 0.935 & 4.234 & 0.214 & {\bf 0.827} & 0.945 & 0.977 \\
    
    +SIGL & \texttt{K} & {\bf 0.139} & {\bf 0.808} & {\bf 4.180} & {\bf 0.209} & 0.826 & {\bf 0.948} & {\bf 0.980} \\
    
    \hline
    Zhou~\cite{zhou2017unsupervised} & \texttt{CS+K} & 0.190 & 1.436 & 4.975 & 0.258 & 0.735 & 0.915 & 0.968 \\
    
    Mahjourian~\cite{mahjourian2018unsupervised} & \texttt{CS+K} & 0.151 & 0.949 & 4.383 & 0.227 & 0.802 & 0.935 & 0.974 \\
    
    Yin~\cite{yin2018geonet} {\tiny \texttt{ResNet}}\tnote{*} & \texttt{CS+K} &  / & / & / & / & / & / & / \\
    
    +SIGL & \texttt{CS+K} &  {\bf 0.141} & {\bf 0.837} & {\bf 4.160} & {\bf 0.209} & {\bf 0.823} & {\bf 0.947} & {\bf 0.980} \\
    
    \hline
    Wang~\cite{wang2017learning} {\tiny \texttt{PoseCNN+DDVO}}\tnote{$\dagger$} & \texttt{CS+K} & 0.142 & 0.901 & 4.202 & 0.213 & 0.827 & 0.946 & 0.978 \\
    
    +SIGL & \texttt{CS+K} & {\bf 0.135} & {\bf 0.832} & {\bf 4.119} & {\bf 0.206} & {\bf 0.836} & {\bf 0.949} & {\bf 0.980} \\
    
    \hline
    \end{tabular}
    \begin{tablenotes}
        \item [*] Not available.
        \item [$\dagger$] Evaluated with prediction released by the author.
    \item +SIGL: training with SIGL enabled
    \end{tablenotes}
\end{threeparttable}
\end{scriptsize}
\end{table}

\setlength{\tabcolsep}{1pt}

\subsection{Ablation study}
\label{sect-ablation}
To study the contribution of each semantic category to the performance improvement, we performed an ablation study: We apply our losses to different semantic categories, one at a time, train the network until convergence, and show how the quality of depth prediction varies (Table~\ref{tab-ablation}). In Table~\ref{tab-ablation}, Godard \textit{et al}. \cite{godard2017unsupervised} is the baseline model where only the most generic regularizers, {\em e.g.,} smoothness and consistency, are used. The second column indicates the semantic category of which the depth prediction is regularized using our losses in addition to the generic regularizers. For the meaning of the semantic categories, see Sect.~\ref{sect-pspnet}.

It turns out that the ``flat'' category contributes most to the performance gain over the baseline model, which is expected because most of the KITTI images contain a large portion of roads and sidewalks. We also observed that regularization of the ``construction'' and ``vehicle'' category provides reasonable improvement while the ``nature'' category (trees and hedges) helps a little. Applying our priors to the ``human'', ``sky'' and ``object'' categories does not consistently improve over the baseline, for the following reasons: ``sky'' does not have well-defined surface normals; ``human'' has deformable surfaces of which normals can point arbitrarily; ``object'' category consists of thin structures which project to few pixels rendering it hard to apply segmentation and our losses. The best is achieved when we apply our losses to ``vehicle'', ``construction'' and ``flat'' categories, denoted by \texttt{V+C+F} in Table~\ref{tab-ablation}.

\setlength{\tabcolsep}{2pt}
\begin{table}[h]
\caption{Ablation study on KITTI.}
\label{tab-ablation}
\centering
\begin{scriptsize}
    \begin{tabular}{l|l|cccc|ccc}
    \hline 
    Method & Category & \multicolumn{4}{c|}{Error metric} & \multicolumn{3}{c}{Accuracy $(\delta < )$ } \\ 
        & & {\tiny AbsRel} & {\tiny SqRel} & {\tiny RMSE} & {\tiny RMSElog} & $1.25$ & $1.25^2$ & $1.25^3$ \\
    \hline \hline
    Godard~\cite{godard2017unsupervised} & \texttt{/} & 0.148 & 1.344 & 5.927 & 0.247 & 0.803 & 0.922 & 0.964\\
    \hline
    Ours & {\scriptsize \texttt{Human}} & 0.152 & 1.394 & 5.945 & 0.251 & 0.801 & 0.921 & 0.963 \\
    Ours & {\scriptsize \texttt{Sky}} & 0.148 & 1.368 & 5.864 & 0.245 & 0.807 & 0.923 & 0.964 \\
    Ours & {\scriptsize \texttt{Object}} & 0.146 & 1.335 & 5.986 & 0.249 & 0.800 & 0.920 & 0.963 \\
    Ours & {\scriptsize \texttt{Nature}} & 0.146 & 1.292 & 5.826 &  0.247 &  0.804 & 0.923 & 0.964 \\
    \hline
    Ours & {\scriptsize \texttt{Vehicle}} & 0.143 & 1.304 & 5.797 &  0.241 &  0.814 & 0.927 & 0.966 \\
    Ours & {\scriptsize \texttt{Construction}} & 0.142 & 1.252 & 5.729 &  0.240 &  0.810 & 0.928 & 0.967 \\
    Ours & {\scriptsize \texttt{Flat}} & 0.141 & 1.270 & 5.779 & 0.239 & 0.814 & 0.927 & 0.966 \\
    \hline
    Ours & \texttt{V+C+F} & 0.139 & 1.211 & 5.702 &  0.239 &  0.816 &  0.928 & 0.966 \\
    \hline
    \end{tabular}
\end{scriptsize}
\end{table}

\subsection{Generalize to other datasets: Make3D}
\label{sect-make3d}
To showcase the generalizability of our approach, we follow the convention of \cite{godard2017unsupervised,zhou2017unsupervised,yin2018geonet,wang2017learning}: Our model trained \textit{only} on KITTI Eigen split is directly tested on Make3D~\cite{saxena2009make3d}.  Make3D contains 534 images with $2272 \times 1707$ resolution, of which 134 are used for testing.\footnote{Ideally we want to test on the whole Make3D dataset since we do not train on Make3D, but other methods to which we compare train on it. For a fair comparison, we only use the 134 images for testing.} Low resolution ground truth depths are given as $305 \times 55$ range maps and must be resized and interpolated for evaluation. We follow \cite{godard2017unsupervised} and \cite{zhou2017unsupervised} in applying a central cropping to generate a $852 \times 1707$ crop centered on the image. We use the standard $C1$ evaluation metrics for Make3D and measure our performance on depths less than 70 meters. Table~\ref{tab-make3d} shows a quantitative comparison to the competitors, both supervised and self-supervised, with two different training settings. Note that the results of \cite{karsch2012depth,liu2016learning,laina2016deeper} are directly taken from \cite{godard2017unsupervised}. Since the exact cropping scheme used in \cite{godard2017unsupervised} is not available, we re-implemented it closely following the description in \cite{godard2017unsupervised}. We trained our model on KITTI Eigen split and compared against models of \cite{godard2017unsupervised,zhou2017unsupervised,yin2018geonet,wang2017learning} also trained on Eigen split (as provided by the authors) for a fair comparison.

\begin{figure}
\centering
\begin{tabular}{ccc}
    \includegraphics[width=0.30\linewidth]{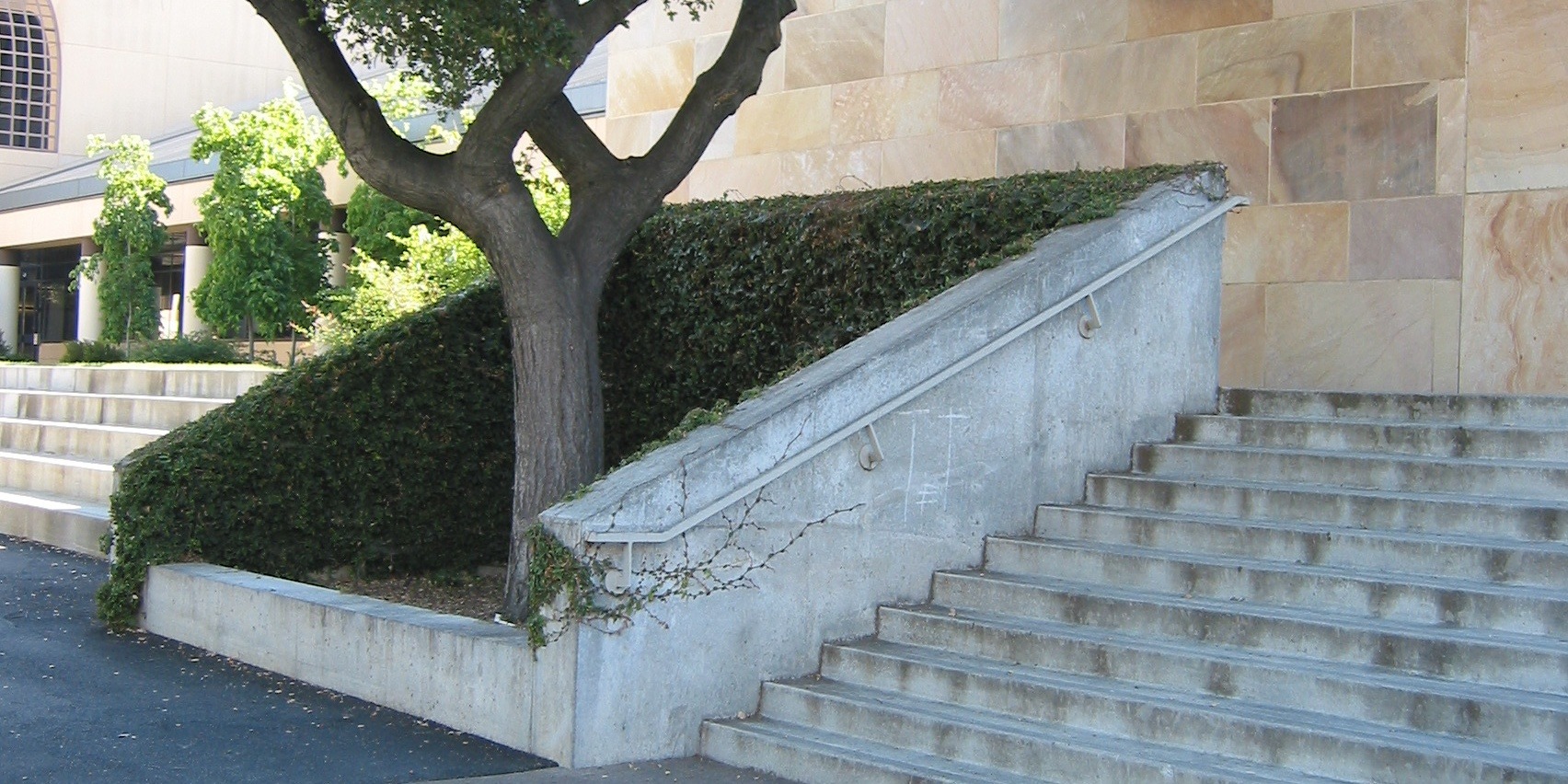} &
    \includegraphics[width=0.30\linewidth]{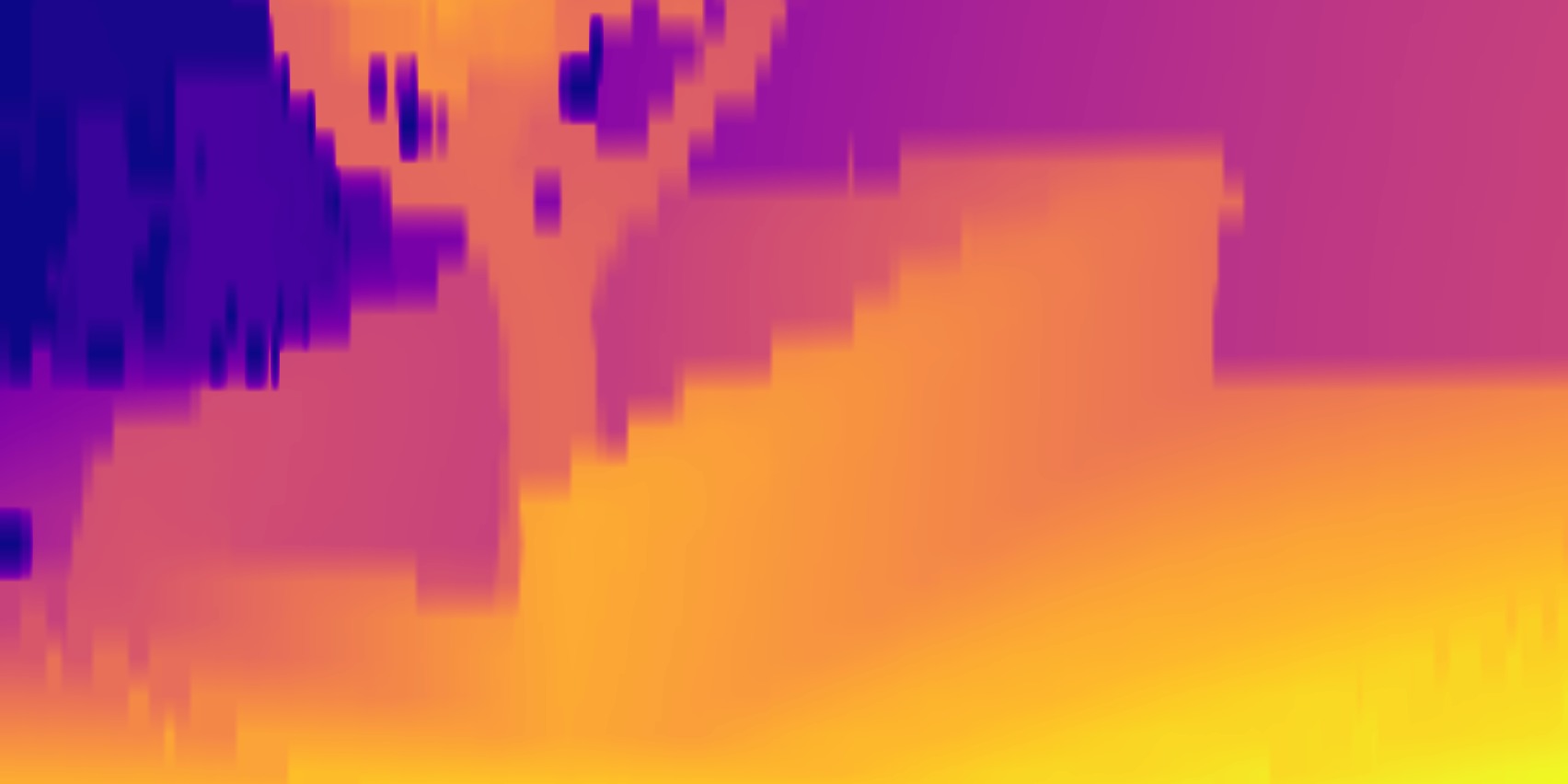} &
    \includegraphics[width=0.30\linewidth]{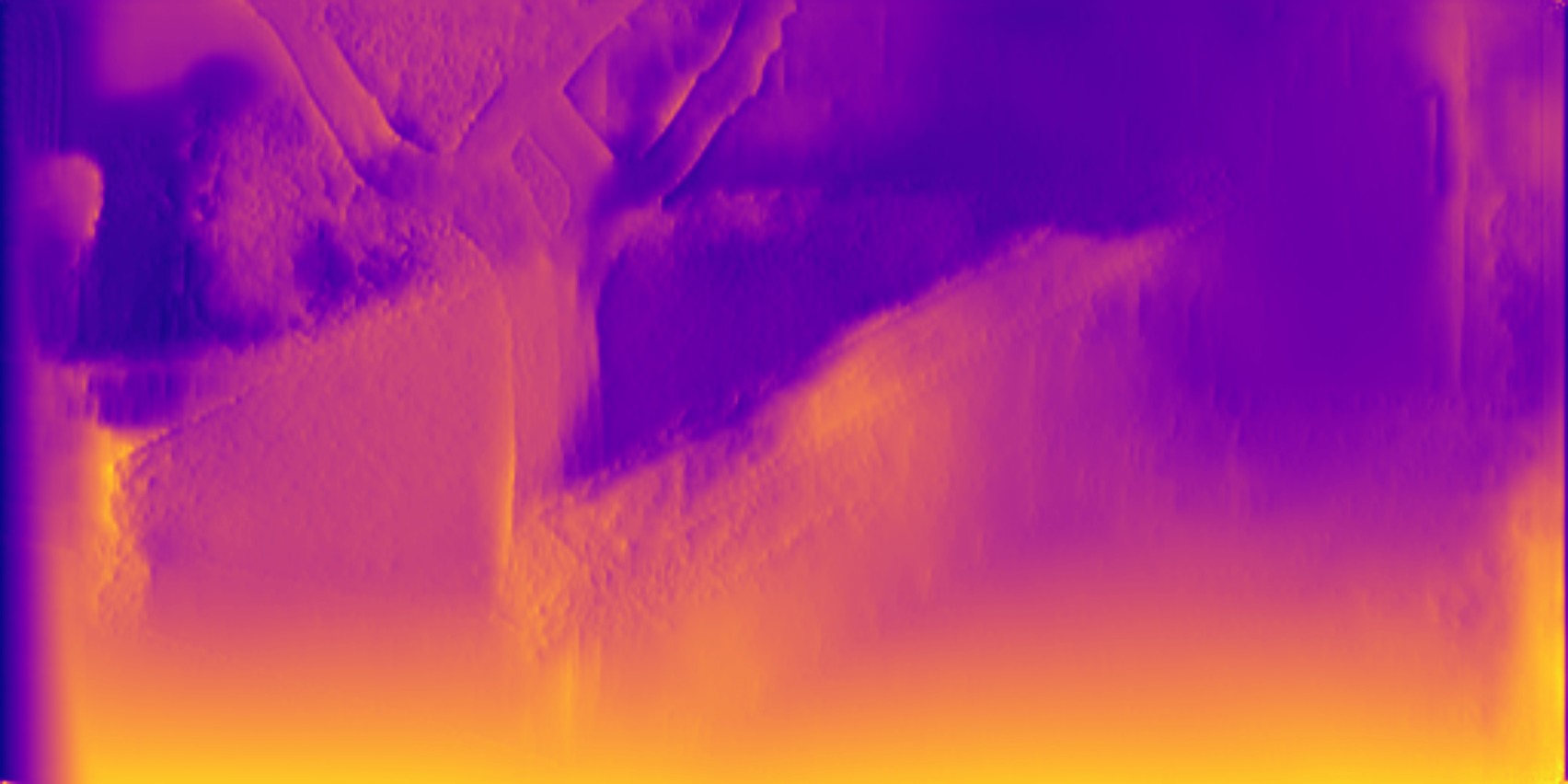} \\
    \includegraphics[width=0.30\linewidth]{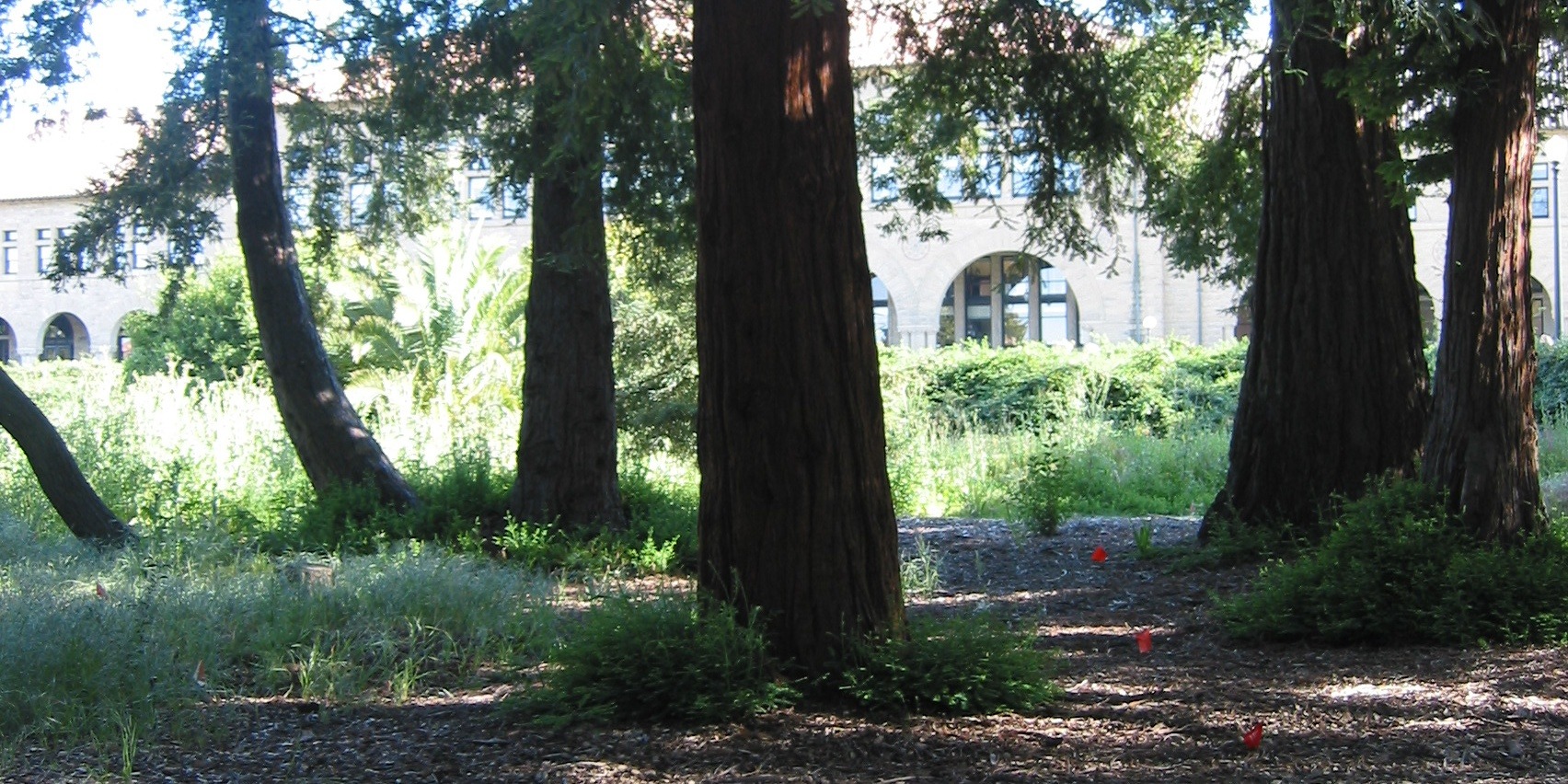} &
    \includegraphics[width=0.30\linewidth]{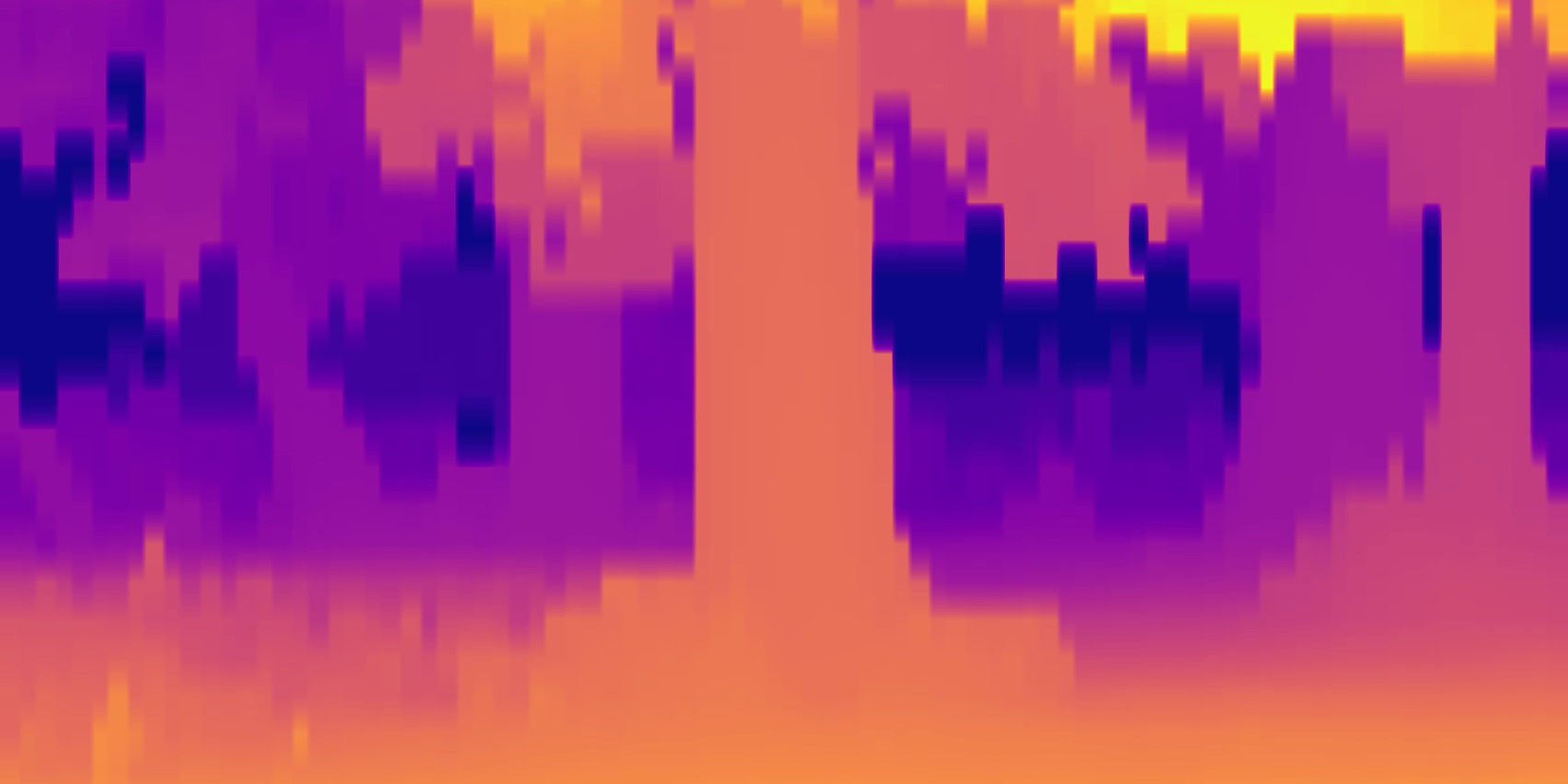} &
    \includegraphics[width=0.30\linewidth]{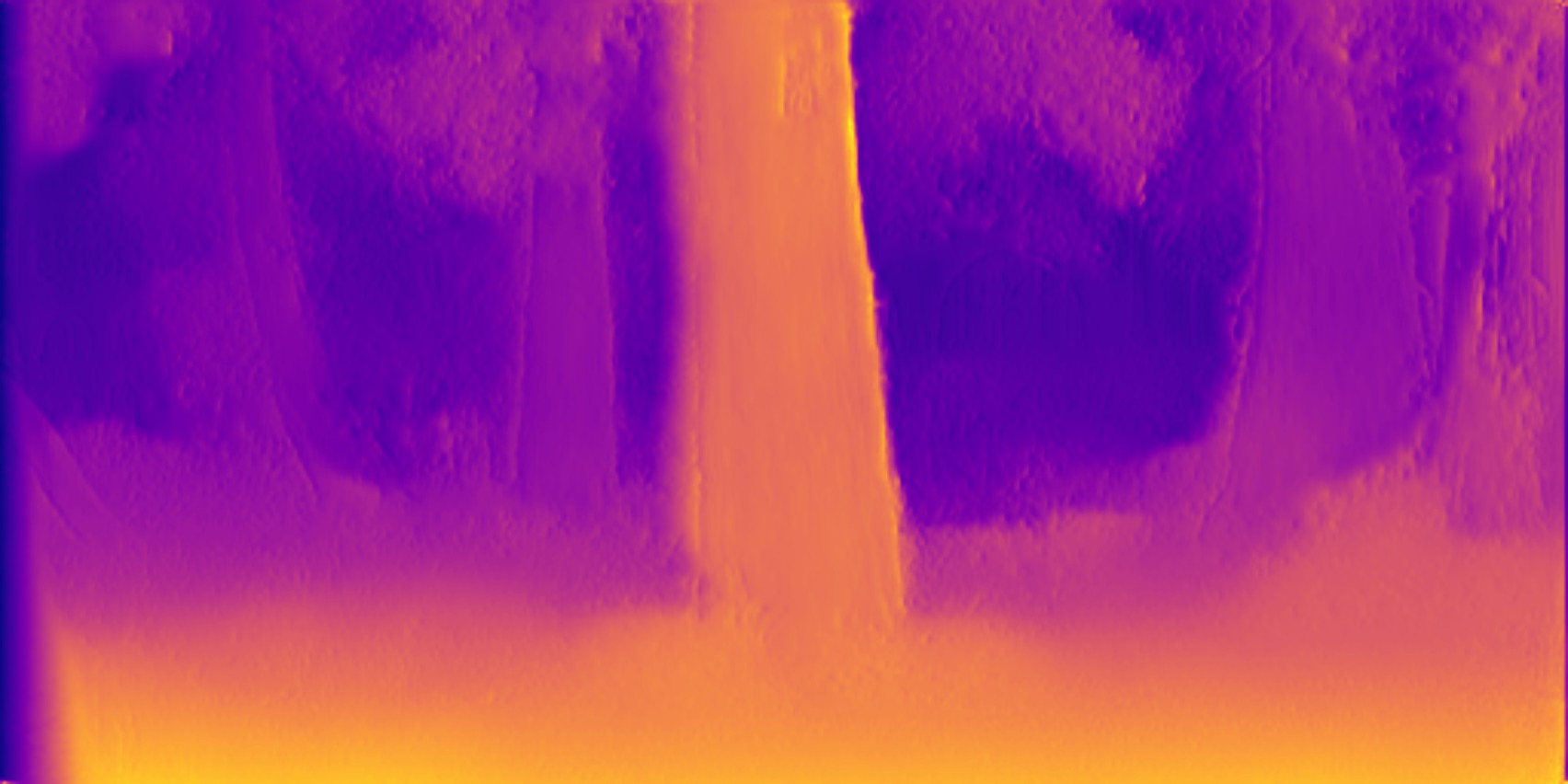} \\
    \includegraphics[width=0.30\linewidth]{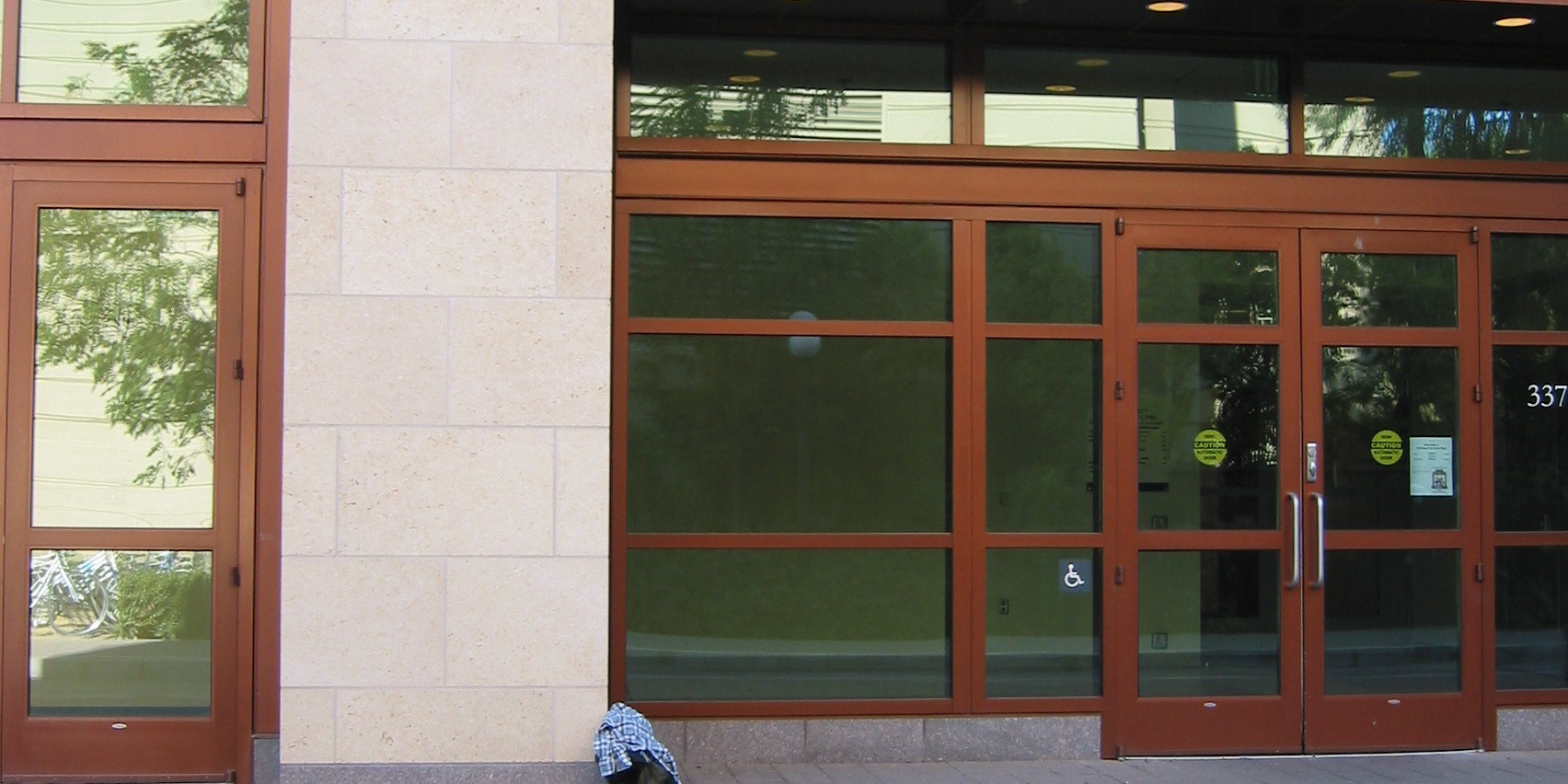} &
    \includegraphics[width=0.30\linewidth]{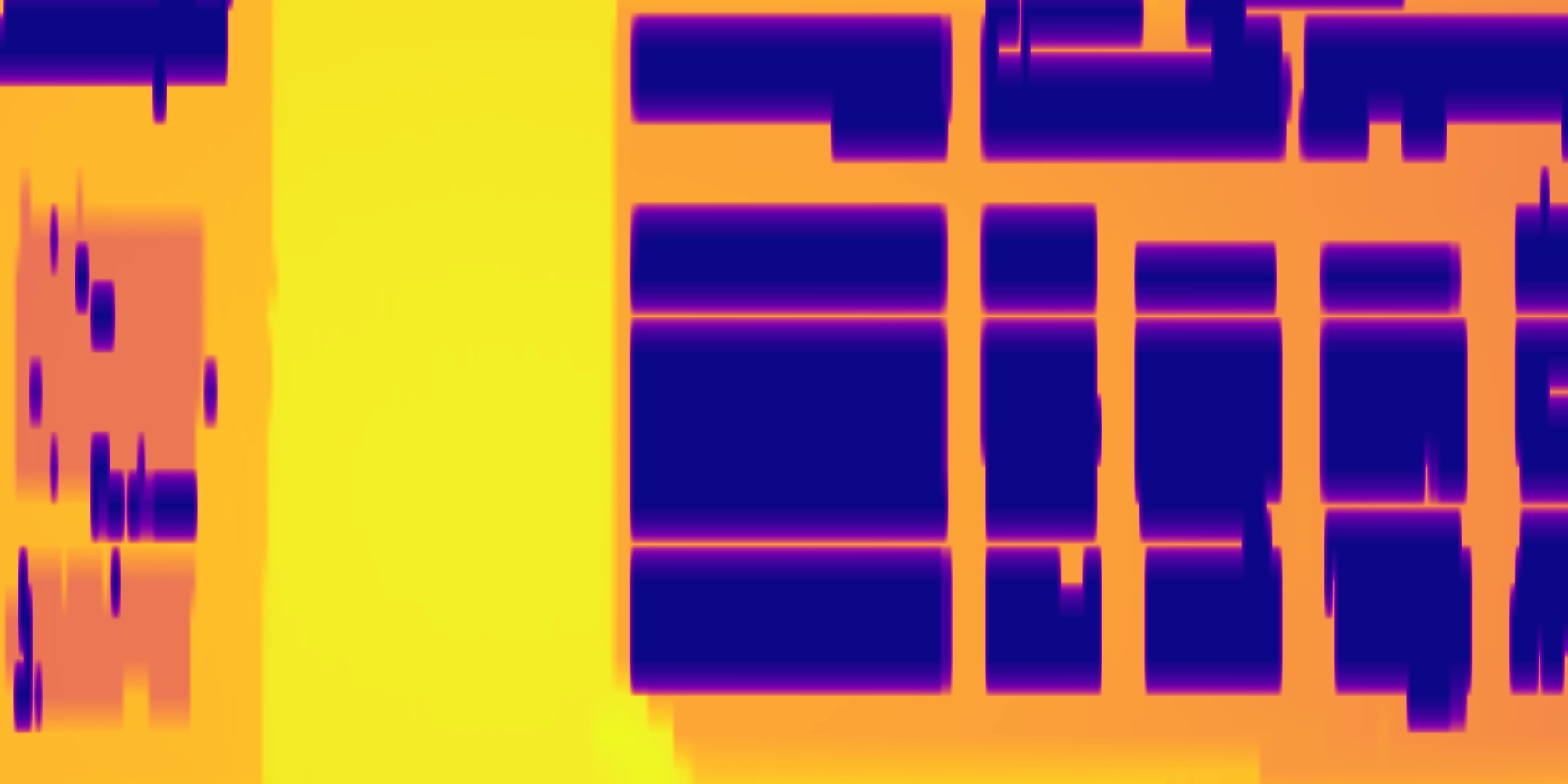} &
    \includegraphics[width=0.30\linewidth]{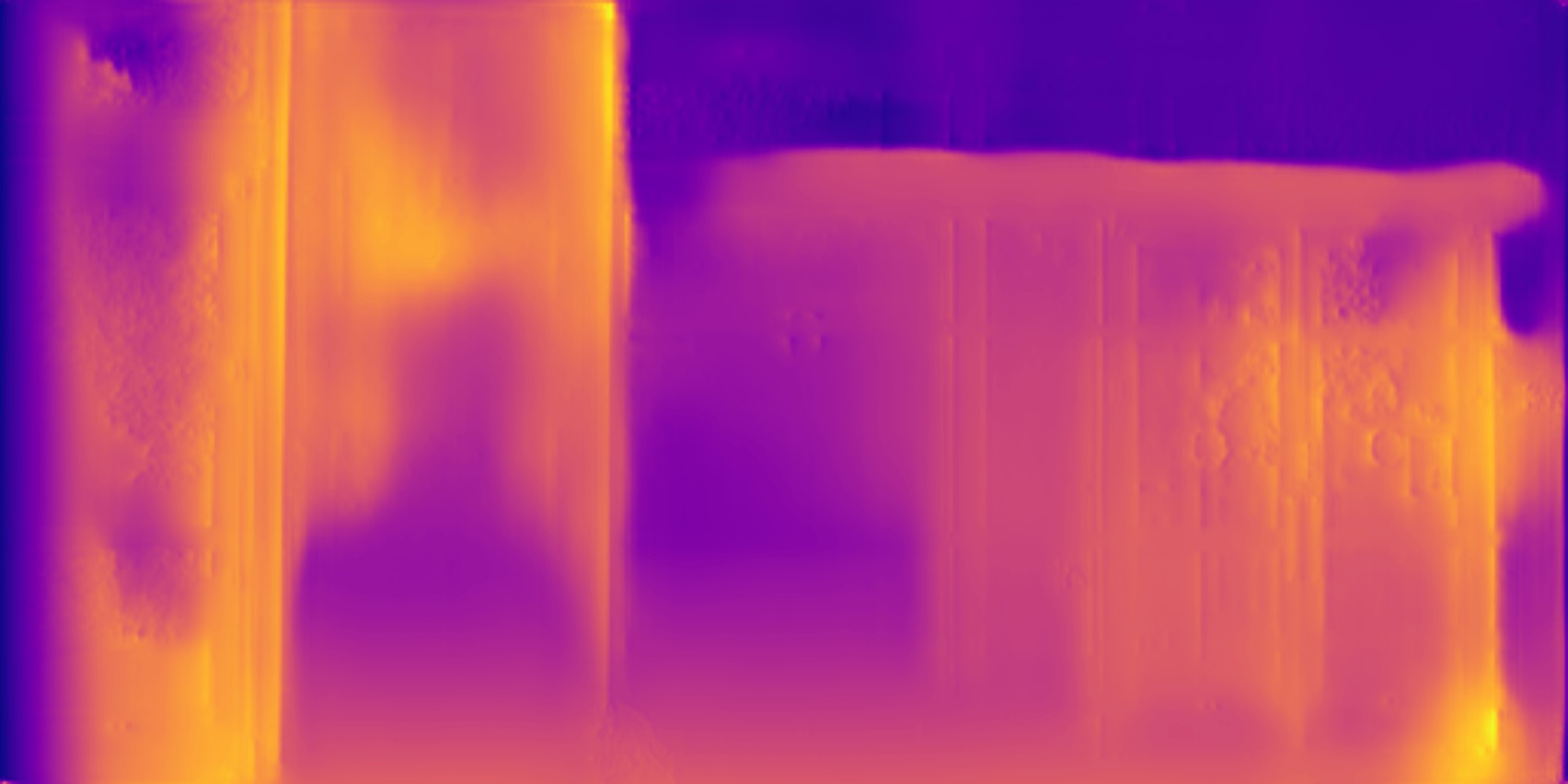}\\
\end{tabular}
 \caption{\small \textit{Qualitative results on Make3D.} Left to right, each row shows an input RGB image, the corresponding ground truth disparity map and our prediction. Our model is {\it only} trained on KITTI and directly applied to Make3D.}
 \label{fig-make3d}
 \vspace{-15pt}
\end{figure}

A careful inspection of the baseline models (Godard~\cite{godard2017unsupervised}  in stereo and Yin~\cite{yin2018geonet} in monocular supervision) versus ours reveals that the application of our losses does not hurt the generalizability of the baselines. Fig.~\ref{fig-make3d} shows some qualitative results on Make3D. 
Though our model registers some failure cases in texture-less regions, a rough scene layout is present in the prediction. Regarding that the model is only trained on KITTI, of which the data modality is very different from that of Make3D, the prediction is sensible. But after all, a single image only affords to hypothesize depth, so we expect that any method using such predictions would have mechanisms to handle model deficiencies.

\setlength{\tabcolsep}{3pt}
\begin{table}
\caption{Generalizability test on Make3D.}
\label{tab-make3d}
\centering
\begin{scriptsize}
\begin{threeparttable}[c]
    \begin{tabular}{l|c|cccc}
    \hline 
    Method & Supervision & AbsRel & SqRel & RMSE & $\log_{10}$ \\
    \hline \hline
    TrainSetMean & Depth & 0.893 & 15.517 & 11.542 & 0.223 \\
    Karsch~\cite{karsch2012depth} & Depth & 0.417 & 4.894 & 8.172 & 0.144 \\
    Liu~\cite{liu2016learning} & Depth & 0.462 & 6.625 & 9.972 & 0.161 \\
    Laina~\cite{laina2016deeper} & Depth & {\bf 0.198} & {\bf 1.665} & {\bf 5.461} & {\bf 0.082} \\
    \hline \hline
    Godard~\cite{godard2017unsupervised} {\tiny \texttt{VGG}} & Stereo & 0.468 & 9.236 & 12.525 & 0.165 \\
    \textbf{Ours} & Stereo & {\bf 0.458} & {\bf 8.681} & {\bf 12.335} & {\bf 0.164} \\
    \hline \hline
    Zhou~\cite{zhou2017unsupervised} & Mono & 0.407 & 5.367 & 11.011 & 0.167 \\
    Yin~\cite{yin2018geonet}{\tiny \texttt{ResNet}} & Mono & 0.376 & 4.645 & 10.350 & 0.152 \\
    Wang~\cite{wang2017learning}{\tiny \texttt{PoseCNN+DDVO}}  & Mono & 0.387 & 4.720 & {\bf 8.09} & 0.204 \\
    {\bf Ours}  & Mono & {\bf 0.356} & {\bf 4.517} & 10.047 & {\bf 0.144} \\
    \hline
    \end{tabular}
    \begin{tablenotes}
    \end{tablenotes}
\end{threeparttable}
\end{scriptsize}
\end{table}

\setlength{\tabcolsep}{1pt}

\subsection{Evaluation on indoor datasets}
\label{sect-visma}
To the best of our knowledge, none of the top-performing methods in self-supervised depth prediction have shown experimental results beyond planar motion, {\em i.e.,} driving scenarios such as KITTI and CityScapes, probably due to two reasons: Lack of rectified stereo pairs for training  (\cite{godard2017unsupervised,zhan2018unsupervised}) and difficulty to learn complex ego-motion along with depth prediction from video sequences (\cite{zhou2017unsupervised,yin2018geonet,wang2017learning}).

However, with two modifications to the \variant{GeoNet} model of Yin~\cite{yin2018geonet} -- a multi-task learning approach where ego-motion and depth prediction are jointly learned, we managed to train our model and outperform \variant{GeoNet} on publicly available VISMA~\cite{fei2018visual} dataset which features monocular videos of indoor scenes captured by a hand-held visual-inertial sensor platform under challenging motion. As a first modification, we replace the pose network in \variant{GeoNet} with pose estimation from a VIO system~\cite{tsotsos2015robust}, which makes the network easier to train (we call this model \variant{OursVIO}). Second, to further improve the quality of predicted depth maps, we impose our gravity-induced regularization terms to \variant{OursVIO}, where gravity is also estimated online by VIO. Our second model is named \variant{OursVIO++}.

\begin{figure}
\centering
\begin{tabular}{ccc}
    \includegraphics[width=0.30\linewidth]{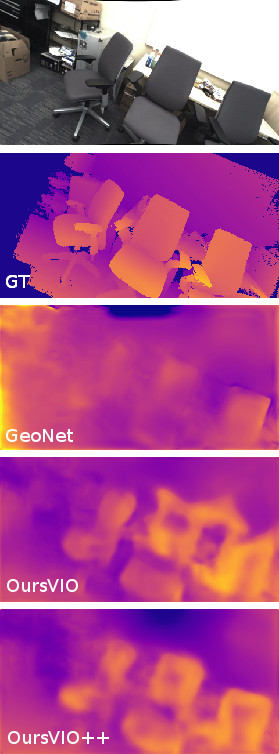} &
    \includegraphics[width=0.30\linewidth]{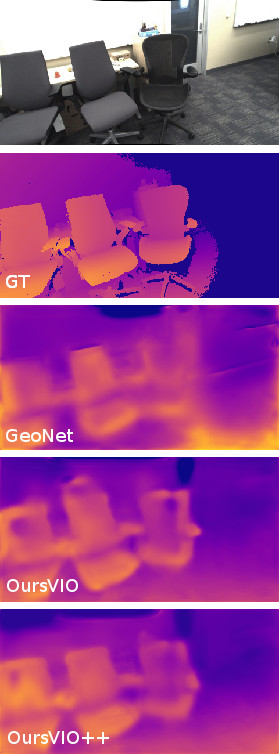} &
    \includegraphics[width=0.30\linewidth]{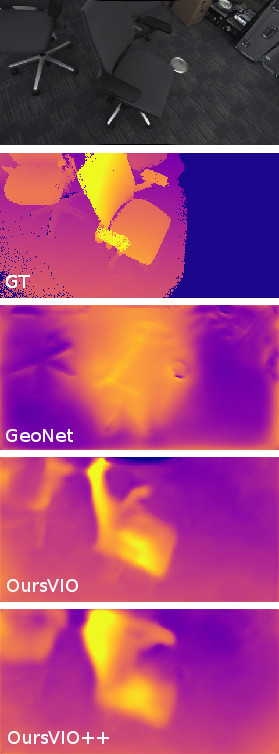}
\end{tabular}
\caption{\small \textit{Qualitative comparison on VISMA validation.} Top to bottom, each column shows an input RGB image, the corresponding ground truth inverse depth map, results of \variant{GeoNet} (baseline), \variant{OursVIO}, and \variant{OursVIO++}. Both \variant{OursVIO} and \variant{OursVIO++} show largely improved results over the baseline, especially for images captured at extreme viewpoint (large in-plane rotation and top-down view). \variant{OursVIO++} (with gravity-induced priors) further improves over \variant{OursVIO} (without priors) at planar regions, {\em e.g.,} the chair backs, where holes have been filled.}
\label{fig-visma}
\end{figure}

VISMA dataset contains time-stamped monocular videos ($30$ Hz) from a PointGrey camera and inertial measurements ($100$ Hz) from an Xsens unit, which are used in both VIO and network training. RGB-D reconstructions (dense point clouds) of the same scenes from a Kinect are also available, along with the spatial alignment $g_{\text{VIO}\leftarrow\text{RGBD}} \in \se$ from RGB-D to VIO provided by the author. To get ground truth depth for cross-modality validation, we apply $g_{\text{VIO}\leftarrow\text{RGBD}}$ to the dense point clouds which are then projected to the PointGrey video frames. PSPNet trained on ADE20K~\cite{zhou2017scene} produces segmentation masks for training.\footnote{Among the 91 categories in ADE20K which PSPNet is trained on, we select ``floor'', ``ceiling'', ``wall'', ``window'', ``door'', ``building'', ``chair'', ``cabinet'', ``desk'', ``table'' to apply our losses.} 
Of the $10K$ frames in VISMA, we remove static ones and construct 3-frame sequences (triplet) which are five frames apart in the original video to ensure sufficient parallax, resulting $8,511$ triplets in total. We randomly sample $100$ triplets for validation and use the rest for training.  
Fig.~\ref{fig-visma} and Table~\ref{tab-visma} show comparisons of \variant{GeoNet}, \variant{OursVIO} and \variant{OursVIO++}, all trained from scratch on VISMA until validation error stops decreasing. Both \variant{OursVIO} and \variant{OursVIO++} improve over the baseline model by a large margin. Moreover, \variant{OursVIO++} trained with our gravity-induced losses has the capability to further refine results of \variant{OursVIO} trained without our losses.

\setlength{\tabcolsep}{3pt}
\begin{table}
\caption{Quantitative results on VISMA validation.}
\label{tab-visma}
\centering
\begin{threeparttable}[c]
    \begin{tabular}{l|cccc|ccc}
    \hline 
    Method & \multicolumn{4}{c|}{Error metric} & \multicolumn{3}{c}{Accuracy $(\delta < )$ } \\ 
        & {\scriptsize AbsRel} & {\scriptsize SqRel} & {\scriptsize RMSE} & {\scriptsize RMSElog} & $1.25$ & $1.25^2$ & $1.25^3$ \\
    \hline \hline
    \texttt{GeoNet} & 0.204 & 0.157 & 0.518 & 0.250 & 0.702 & 0.914 & 0.975 \\
    \hline
    \texttt{OursVIO} & 0.154 & 0.111 & 0.446 & 0.211 & 0.796 & 0.940 & 0.983 \\
    \hline
    \texttt{OursVIO++} & {\bf 0.149} & {\bf 0.105} & {\bf 0.421} & {\bf 0.202} & {\bf 0.820} & {\bf 0.947} & 0.983 \\
    \hline
    \end{tabular}
    \begin{tablenotes}
    \end{tablenotes}
\end{threeparttable}
\end{table}

\setlength{\tabcolsep}{1.5pt}

\section{\textsc{Discussion}}
Gravity informs the shape of objects populating the scene, which is a powerful prior to visual scene analysis. We have presented a simple illustration of this power by adding a prior to standard monocular depth prediction methods that biases the normals of surfaces of known classes to align to gravity or its complement. Far more can be done: 
While in this work we use known biases in the shape of certain object classes, such as the fact that roads tend to be perpendicular to gravity, in the future we could learn such biases directly.


\end{document}